\PassOptionsToPackage{unicode}{hyperref}
\PassOptionsToPackage{hyphens}{url}
\documentclass[
  11pt,
]{article}
\usepackage{amsmath,amssymb}
\usepackage{iftex}
\ifPDFTeX
  \usepackage[T1]{fontenc}
  \usepackage[utf8]{inputenc}
  \usepackage{textcomp} 
\else 
  \usepackage{unicode-math} 
  \defaultfontfeatures{Scale=MatchLowercase}
  \defaultfontfeatures[\rmfamily]{Ligatures=TeX,Scale=1}
\fi
\usepackage{lmodern}
\usepackage{newunicodechar}
\newunicodechar{−}{\ensuremath{-}}
\newunicodechar{Δ}{\ensuremath{\Delta}}
\newunicodechar{≈}{\ensuremath{\approx}}
\newunicodechar{≥}{\ensuremath{\geq}}
\newunicodechar{→}{\ensuremath{\rightarrow}}
\newunicodechar{×}{\ensuremath{\times}}
\newunicodechar{±}{\ensuremath{\pm}}
\newunicodechar{⁻}{\textsuperscript{-}}
\newunicodechar{⁷}{\textsuperscript{7}}
\newunicodechar{¹}{\textsuperscript{1}}
\newunicodechar{·}{\ensuremath{\cdot}}
\newunicodechar{³}{\textsuperscript{3}}
\newunicodechar{²}{\textsuperscript{2}}
\newunicodechar{⁶}{\textsuperscript{6}}
\newunicodechar{⁸}{\textsuperscript{8}}
\newunicodechar{ρ}{\ensuremath{\rho}}
\newunicodechar{ε}{\ensuremath{\varepsilon}}
\newunicodechar{τ}{\ensuremath{\tau}}
\newunicodechar{°}{\ensuremath{^\circ}}
\newunicodechar{≤}{\ensuremath{\leq}}
\newunicodechar{≠}{\ensuremath{\neq}}
\newunicodechar{∈}{\ensuremath{\in}}
\newunicodechar{↔}{\ensuremath{\leftrightarrow}}
\newunicodechar{‖}{\ensuremath{\|}}
\newunicodechar{√}{\ensuremath{\sqrt{}}}
\newunicodechar{⌊}{\ensuremath{\lfloor}}
\newunicodechar{⌋}{\ensuremath{\rfloor}}
\newunicodechar{λ}{\ensuremath{\lambda}}
\newunicodechar{α}{\ensuremath{\alpha}}
\newunicodechar{η}{\ensuremath{\eta}}
\newunicodechar{σ}{\ensuremath{\sigma}}
\newunicodechar{⊇}{\ensuremath{\supseteq}}
\newunicodechar{⁰}{\textsuperscript{0}}
\newunicodechar{⁴}{\textsuperscript{4}}
\newunicodechar{⁹}{\textsuperscript{9}}
\newunicodechar{…}{\dots}
\newunicodechar{¶}{\P}
\newunicodechar{÷}{\ensuremath{\div}}
\usepackage{float}
\usepackage{caption}
\ifPDFTeX\else
\fi
\IfFileExists{upquote.sty}{\usepackage{upquote}}{}
\IfFileExists{microtype.sty}{
  \usepackage[]{microtype}
  \UseMicrotypeSet[protrusion]{basicmath} 
}{}
\makeatletter
\@ifundefined{KOMAClassName}{
  \IfFileExists{parskip.sty}{%
    \usepackage{parskip}
  }{
    \setlength{\parindent}{0pt}
    \setlength{\parskip}{6pt plus 2pt minus 1pt}}
}{
  \KOMAoptions{parskip=half}}
\makeatother
\usepackage{xcolor}
\usepackage[margin=1in]{geometry}
\usepackage{longtable,booktabs,array}
\usepackage{calc} 
\usepackage{etoolbox}
\makeatletter
\patchcmd\longtable{\par}{\if@noskipsec\mbox{}\fi\par}{}{}
\makeatother
\IfFileExists{footnotehyper.sty}{\usepackage{footnotehyper}}{\usepackage{footnote}}
\makesavenoteenv{longtable}
\usepackage{graphicx}
\makeatletter
\def\maxwidth{\ifdim\Gin@nat@width>\linewidth\linewidth\else\Gin@nat@width\fi}
\def\maxheight{\ifdim\Gin@nat@height>\textheight\textheight\else\Gin@nat@height\fi}
\makeatother
\setkeys{Gin}{width=\maxwidth,height=\maxheight,keepaspectratio}
\makeatletter
\def\fps@figure{htbp}
\makeatother
\providecommand{\tightlist}{%
  \setlength{\itemsep}{0pt}\setlength{\parskip}{0pt}}
\ifLuaTeX
  \usepackage{selnolig}  
\fi
\usepackage{bookmark}
\IfFileExists{xurl.sty}{\usepackage{xurl}}{} 
\hypersetup{
  hidelinks,
  pdfcreator={LaTeX via pandoc}}

\author{}
\date{}

\begin{document}

\section{Geometric Evolution Maps: Extracting Stable Concept Probes from
Transformer Residual
Streams}\label{geometric-evolution-maps-extracting-stable-concept-probes-from-transformer-residual-streams}

\textbf{James Henry}\\
\emph{Independent Researcher} · jamesrahenry@henrynet.ca ·
\href{https://orcid.org/0009-0005-7126-9466}{ORCID 0009-0005-7126-9466}

Version 1: May 25, 2026 · \textbf{Version 2: August 12, 2026}

\begin{center}\rule{0.5\linewidth}{0.5pt}\end{center}

\textbf{Version 2.} This paper supersedes v1 (arXiv:2605.25848). It is a
substantial revision: the corpus grew from 23 to 29 base models, several
v1 claims are withdrawn, and one control is replaced by a corrected
version that reverses its conclusion. A reader holding v1 should read
\textbf{Appendix E} before relying on any number from it; readers new to
this work need nothing from there. Full edit-level provenance is in
\texttt{V2\_CHANGELIST.md} in the accompanying repository.

\begin{center}\rule{0.5\linewidth}{0.5pt}\end{center}

\subsection{Abstract}\label{abstract}

A concept probe is only as reliable as the layer it is taken from.
Probing at a fixed late layer, or at the peak of a separation curve,
ignores a structural feature of how concepts form: the probe direction
rotates substantially during assembly and does not settle until after
the Concept Allocation Zone (CAZ) {[}Henry, 2026a{]} in which it forms.

We introduce \textbf{Geometric Evolution Maps (GEMs)}. A GEM records a
concept's directional trajectory across one CAZ segment, takes the
settled direction at that segment's final layer, and locates the
\emph{handoff layer} immediately beyond it --- the first layer at which
that direction can be evaluated outside the window it was estimated on.
The handoff layer is derived from the segment boundary, not detected by
a rotation criterion. A concept typically occupies several segments; we
call that set its \emph{atlas}.

This rotation is substantial and consistent across a wide range of
architectures and concept types --- a property of the representation
itself, not an artifact of how it is estimated. The extracted direction
is causal, not an artifact of projection: ablating it suppresses far
more separation than ablating a random direction. But no single site
carries that effect alone. Probing after the rotation completes is more
precise than probing during it, though a depth-matched control shows the
advantage comes from probing in later layers, not any privilege attached
to the handoff boundary specifically --- the handoff layer is simply
where, in this data, the direction is observed to stop rotating --- and
within a concept's atlas, no single GEM accounts for the whole signal
either. A small number of structured exceptions are documented and
explained rather than left unaccounted for.

A concept's atlas --- its full set of GEMs, not any single one of them
--- is therefore the unit this method delivers for downstream use. GEM
provides the ablation-targeting methodology used by the CAZ Validation
paper {[}Henry, 2026c{]} in this program.

\subsection{1. Introduction}\label{introduction}

A linear probe of a transformer's residual stream is most reliable at a
layer where the concept has been fully assembled and stably encoded. The
central question this paper addresses is: \emph{which layer is that?}

Three current answers appear in the literature. The first is the final
layer, on the grounds that information must be maximally processed
there. The second is the peak of a separation score function --- the
layer where a contrastive probe scores highest. The third is a fixed
middle-to-late depth selected by hyperparameter search or convention.

Each of these answers focuses on selecting a layer, not on whether the
concept's direction is even stable once you get there. In this paper we
show that concept representations in transformer residual streams are
not static between layers: they emerge gradually, rotate substantially
during their allocation zone, and then --- if the concept is well-formed
in that model --- \emph{settle} into a direction that is markedly more
stable than during assembly (§3.2). We call the transition point the
\emph{handoff layer}: the layer at which the concept direction finishes
rotating and locks in.

This paper introduces \textbf{Geometric Evolution Maps (GEMs)}, a method
for recording the directional trajectory of a concept across each of its
CAZ regions {[}Henry, 2026a{]}, taking the settled concept direction at
a region's end as the probe, and evaluating it at the handoff layer
immediately beyond. A concept typically has several such regions and
therefore several GEMs; we call that set its \emph{atlas} (§3.1). Three
findings anchor the validation that follows: concept directions rotate
substantially during assembly and then settle, so probing after settling
completes outperforms probing during rotation --- though the advantage
tracks depth more than the specific settling boundary (§5.5); the
extracted direction is causal rather than an artifact of projection
(§5.3); and no single site is sufficient on its own, whether within a
settled region or across a concept's atlas --- the atlas as a whole is
the unit this method delivers (§5.4).

\subsubsection{1.1 Motivation: Rotation Is
Ubiquitous}\label{motivation-rotation-is-ubiquitous}

Figure 1 summarizes the key motivating observation. Across 493 concept ×
model pairs (29 architectures, 17 concept types), we compute the cosine
similarity between the dominant concept direction at CAZ entry and at
CAZ exit --- the \emph{entry-exit cosine} (EEC). A value of 1.0 would
mean no rotation: the concept direction is identical before and after
the allocation zone. A value of 0.0 would mean the post-assembly
direction is orthogonal to the pre-assembly direction.

\begin{figure}
\centering
\includegraphics{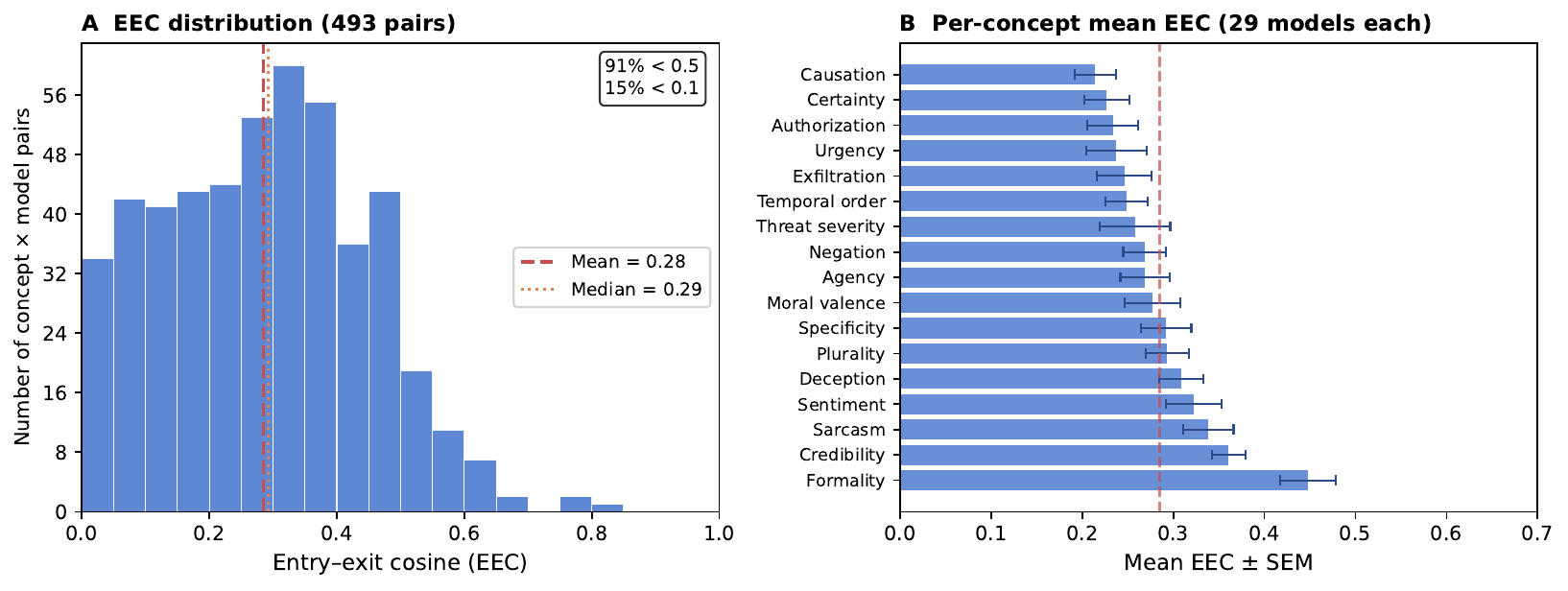}
\caption{Figure 1: Entry-exit cosine (EEC) across 493 concept × model
pairs (29 models × 17 concepts). \textbf{Panel A:} Distribution of EEC
values; mean (0.285, dashed) and median (0.292, dotted) marked.
\textbf{Panel B:} Per-concept mean EEC ± SEM, sorted by mean. Formality
and credibility have the highest per-concept mean EEC (rotate least
within their CAZs). All 17 concepts have mean EEC well below 0.5.}
\end{figure}

The mean EEC across our dataset is \textbf{0.285} (median 0.292). 91.5\%
of pairs have EEC \textless{} 0.5, and 15.4\% are near-orthogonal (EEC
\textless{} 0.1). This is not noise: the maximum rotation per layer
(averaged across pairs) is \textbf{0.380}, and the mean per-GEM rotation
rate is \textbf{0.295}, indicating that single-layer steps within CAZs
can produce large directional shifts. Probing at CAZ entry, or at the
separation peak (which falls inside the CAZ by construction, since the
CAZ is defined to include it), extracts a direction that has not yet
settled. The validation in §5 therefore compares a post-assembly probe
(GEM at \(L_H\)) against an in-assembly probe (peak at
\(L_{\text{peak}}\)) --- not against the best-possible layer, but
against the standard alternative.

\subsubsection{1.2 Contributions}\label{contributions}

\begin{enumerate}
\def\labelenumi{\arabic{enumi}.}
\tightlist
\item
  \textbf{GEM formalization}: a tractable, deterministic record of a
  concept's directional trajectory across one CAZ region --- the
  per-layer direction sequence and its layer window, together with two
  readouts from it: the settled direction (the trajectory's terminal
  reading) and the handoff layer (the first layer past the region
  boundary at which that direction can be evaluated) (§3.1)
\item
  \textbf{Rotation characterization}: concept directions rotate
  substantially during assembly --- mean entry-exit cosine 0.285 across
  493 concept × model pairs (29 architectures, 17 concepts) against a
  measured estimator noise floor of 0.970, so the rotation is a property
  of the representation and not of the estimator (§4.1). This is the
  paper's anchor empirical finding; the handoff-layer results below
  follow from it
\item
  \textbf{Peak-layer extraction underperforms while rotation is active}:
  probing at the separation peak --- inside the CAZ, while the direction
  is still rotating --- is measurably less precise than probing after it
  settles, further evidence that no single fixed layer can be picked
  blindly without tracking the trajectory; the GEM (handoff-layer) probe
  beats the peak-layer probe in 341/493 trials (69.2\%; 29 base models ×
  17 concepts); model-level Wilcoxon W=356, N=29, p=1.34×10⁻³
  (one-sided, significant); no multi-head attention (MHA) /
  grouped-query attention (GQA) cohort effect (Fisher p=0.462) ---
  failures follow neither a clean scale nor a cohort pattern; see §5.1,
  §9.3. Item 4 scopes what drives this advantage
\item
  \textbf{Depth-matched control}: with one depth-matched control layer
  per GEM, so both arms ablate the same number of layers at matched
  depth, the handoff layers win in 263/493 pairs (53.3\%, per-model
  win-rate Wilcoxon \(p = 0.456\)) across the full corpus ---
  indistinguishable from chance, with no residual advantage stable
  across defensible exclusion rosters (§5.5). Item 3's advantage is
  therefore a depth effect, not a settling effect: probing past the CAZ
  is what matters, and the handoff layer is simply where the data shows
  rotation has stopped --- not a privileged extraction site within it
\item
  \textbf{Adaptive width rule}: a simple threshold (\(L_H/N\)
  \textgreater{} 0.85) that improves probe quality in 60/79 triggered
  cases (primary corpus), mean +7.44pp for the improved cases and
  ≈+4.60pp net across all 79 triggered cases including failures
\item
  \textbf{Scale characterization}: GEM structure is present even in 70M
  models, and handoff preference does not follow a clean scale trend ---
  the sub-500M band is bimodal rather than a floor, and depth does not
  separate it either (§5.1)
\item
  \textbf{Direction-specificity control}: concept-direction ablation
  reduces separation by a mean of 66.3\% against 0.07\% for random
  directions, beating all ten random seeds in 100\% of pairs (C=17, 29
  base models, n=472, handoff layer; §5.3). The random baseline sits at
  the \(1/d\) level expected by chance in \(d\) dimensions, so the ratio
  between the two scales with model width and is reported descriptively
  rather than as an effect size
\item
  \textbf{Structured failure modes}: gpt2 (ablation overshoot at
  near-final depth --- shared by gpt2-medium at twice the depth, so a
  GPT-2-family effect rather than a depth effect; §7.3) as architectural
  diagnostic; opt-6.7b reaches 13/17 = 76\% at N=250, a rate that
  shifted with corpus size in an earlier vintage (§7.1)
\end{enumerate}

\subsubsection{1.3 Scope and Relationship to Other
Papers}\label{scope-and-relationship-to-other-papers}

\textbf{Scope at a glance.} The framework paper {[}Henry, 2026a{]}
evaluates its pre-registered predictions at C = 7 concepts on a 35-model
corpus (29 base + 6 instruct); this paper works at C = 17 on the 29 base
models; the validation paper {[}Henry, 2026c{]} evaluates at C = 17 on
its 28-base-model roster. The full coverage-and-rationale table is in
{[}Henry, 2026a{]} §1.

GEM addresses the probe extraction step. It presupposes that CAZs ---
the zones where concept directions are allocated --- have been
identified. The CAZ Framework paper {[}Henry, 2026a{]} provides that
detection methodology. GEM is introduced in §4.4 of the CAZ Framework
paper as the extraction answer; this paper provides the full treatment.

The CAZ Validation paper {[}Henry, 2026c{]} uses GEM's ablation protocol
--- the handoff layer \(L_H\) as the ablation target (§3.2) --- for its
cohort ablation experiments (its §6.4--6.5); concept directions in that
paper are centroid differences at each layer, not GEM settled-direction
probes stored and reused. GEM probes are validated as more geometrically
precise (§5.1). GEM is methodologically upstream of the CAZ Validation
paper, for which it provides the ablation targeting methodology.

\begin{center}\rule{0.5\linewidth}{0.5pt}\end{center}

\subsection{2. Background}\label{background}

\subsubsection{2.1 Probing Transformer
Representations}\label{probing-transformer-representations}

Linear probing in transformer representations is well-established
(Conneau et al., 2018; Tenney et al., 2019; Belinkov, 2022). The typical
approach trains a linear classifier on frozen layer activations and
interprets high accuracy as evidence that the concept is encoded at that
layer.

Two methodological concerns motivate our work. First, layer selection in
probing is often ad hoc. Many studies probe the final layer, cite the
token representations ``most relevant'' to the task, or sweep a range
and report the best. This conflates \emph{where a concept is stored}
with \emph{where a probe can best recover it} --- the latter may simply
reflect where the contrastive pairs (matched sentences that express a
concept and sentences that do not; construction detailed in §3.5) most
differ in surface form.

Second, even studies that probe multiple layers typically extract a
\emph{single} direction (the probe weight vector) and treat it as the
concept representation. If the concept direction changes substantially
across layers, a single probe captures only a slice of that trajectory.

GEM addresses both concerns by computing the \emph{trajectory} of the
concept direction and selecting the extraction layer based on when that
trajectory settles.

\subsubsection{2.2 Concept Allocation
Zones}\label{concept-allocation-zones}

The CAZ Framework {[}Henry, 2026a{]} characterizes the regions of depth
within which a concept's separation peaks --- scored saddle-to-saddle
segments of \(S(l)\) that partition the full depth. CAZs are detected
via Fisher-normalized separation score \(S(l)\): the detector locates
local peaks of \(S(l)\) subject to a prominence floor, places region
boundaries at the \textbf{saddle points of \(S(l)\) between consecutive
peaks}, and scores each region by a composite function (prominence ×
coherence boost × \(\sqrt{\text{width}}\)). Velocity \(v(l) = dS/dl\) is
computed and stored per layer but does not enter boundary detection.
Because the segmentation partitions the full depth, a CAZ is
depth-\emph{extended} rather than depth-localized, and the composite
score grades geometric salience rather than membership {[}Henry,
2026a{]}. A CAZ is not the concept itself --- it is a region of the
residual stream in which the model organizes its geometry to serve a
concept.

A critical property of CAZs relevant here: they describe \emph{where} a
concept is allocated, but not the direction of the resulting
representation. The peak separation layer within a CAZ is not
necessarily the layer at which the settled representation is available
--- peak separation may occur while the concept is still actively
rotating between attention sub-structures.

\subsubsection{2.3 Concept Directions and
Rotation}\label{concept-directions-and-rotation}

The dominant direction of a concept at layer \emph{l} is the
difference-in-means direction between positive and negative pairs at
that layer --- the L2-normalized difference of class centroids, defined
formally in §3.1 (not a principal-component decomposition). As \emph{l}
increases from the shallow to deep layers, this direction is not
constant. In models that exhibit CAZ dynamics, the direction typically:

\begin{enumerate}
\def\labelenumi{\arabic{enumi}.}
\tightlist
\item
  Is undefined or noisy before the CAZ onset (near-chance separation)
\item
  Begins to crystallize within the CAZ, showing increasing separation
  and initial directional coherence
\item
  Rotates substantially during the CAZ --- successive layers show high
  angular velocity
\item
  Reaches a stable, high-coherence direction at the region's final layer
  (the settled direction, §3.1)
\item
  Survives, on average, the single step to the handoff layer immediately
  beyond it (§3.2) --- within this GEM's own region, tracking stops
  there, and whether the direction drifts further is not measured; if
  the concept's atlas has a subsequent CAZ region, a new GEM begins
  tracking there, independently estimated rather than a continuation of
  this one (§3.1)
\end{enumerate}

Steps 3--4 are the core phenomenon GEM is designed to detect. Our
analysis shows this rotation is not subtle: the mean entry-exit cosine
across our corpus is 0.285, with maximum per-layer rotation of 0.380.

\begin{center}\rule{0.5\linewidth}{0.5pt}\end{center}

\subsection{3. Geometric Evolution Maps}\label{geometric-evolution-maps}

\subsubsection{3.1 Formal Definition}\label{formal-definition}

Let \(M\) be a transformer with \(N\) layers and hidden dimension \(d\).
For a concept \(c\), let \(\mathcal{D} = \{(x_i^+, x_i^-)\}_{i=1}^K\) be
a set of \(K\) contrastive pairs, where \(x^+\) is a sentence strongly
expressing \(c\) and \(x^-\) is a matched sentence not expressing \(c\)
(see §3.5 for pair construction). Outside this formal definition, this
paper follows the program-wide informal convention of writing ``N=250
pairs'' for corpus size (\(K\) in this notation) rather than layer count
--- the same usage as \texttt{paper\_n250} and the other papers in this
series. Context disambiguates every instance; layer count is otherwise
always \(N\) alone or \(L_H/N\), never ``N='' a number.

At each layer \(l \in \{0, \ldots, N-1\}\), let \(\bar{h}_+^{(l)}\) and
\(\bar{h}_-^{(l)}\) denote the mean residual stream activation over
positive and negative pairs respectively, where \(h_l(x)\) is extracted
at the final token position from the post-MLP residual stream output.

The \textbf{dominant concept direction} at layer \(l\) is the
L2-normalized difference of class centroids:
\[u^{(l)} = \frac{\bar{h}_+^{(l)} - \bar{h}_-^{(l)}}{\left\|\bar{h}_+^{(l)} - \bar{h}_-^{(l)}\right\|}\]

This is the difference-in-means direction (Zou et al., 2023; Marks \&
Tegmark, 2024), oriented from the negative class centroid toward the
positive class centroid.

The \textbf{separation score} at layer \(l\) is the Fisher-normalized
centroid distance between positive and negative class activations:
\[S(l) = \frac{\left\| \bar{h}_+^{(l)} - \bar{h}_-^{(l)} \right\|_2}{\sqrt{\tfrac{1}{2}\left(\operatorname{tr}(\Sigma_+^{(l)}) + \operatorname{tr}(\Sigma_-^{(l)})\right)}}\]

where \(\Sigma_+^{(l)}\), \(\Sigma_-^{(l)}\) are the within-class
covariance matrices at layer \(l\).

The \textbf{angular velocity} at layer \(l > 0\) is:
\[\omega(l) = 1 - \left| u^{(l)} \cdot u^{(l-1)} \right|\]

Equivalently, \textbf{directional stability}
\(DS(l) = \left| u^{(l)} \cdot u^{(l-1)} \right| = 1 - \omega(l)\)
measures the complement: how stable the direction is rather than how
fast it rotates. High \(DS(l)\) near 1.0 means the direction barely
moved; low \(DS(l)\) indicates active rotation. \(\omega(l)\) is
reported in this paper as a rotation \textbf{diagnostic}: it does not
enter the detection of CAZ boundaries or of the handoff layer (§3.2).
\(DS(l)\) is the form used for visualization in the CAZ framework
{[}Henry, 2026a §4.2{]}.

A \textbf{GEM} is the trajectory record of a concept's separating
direction across \textbf{one} CAZ region. For concept \(c\) in model
\(M\) and region \(i\):
\[\text{GEM}(c, M, i) = \left( W_i, \; \left(u^{(\ell)}\right)_{\ell \in W_i}, \; L_{H,i}, \; u_i^{\text{settled}} \right)\]

where \(W_i = [\text{start}_i, \text{end}_i]\) is the region's layer
window, \(\left(u^{(\ell)}\right)_{\ell \in W_i}\) is the sequence of
per-layer dominant directions across it, \(L_{H,i}\) is the
\textbf{handoff layer} (defined below), and
\(u_i^{\text{settled}} = u^{(\text{end}_i)}\) is the \textbf{settled
direction} --- the terminal reading of the trajectory, and the concept
probe this paper evaluates.

A concept typically has several CAZ regions and therefore several GEMs
--- a mean of 2.21 per concept per model in this corpus (§4.1) --- and
we call that set the concept's \textbf{atlas}: each GEM charts one
region, and because the segmentation partitions the full depth (§2.2)
the atlas covers the concept across the whole network. The term is used
in its ordinary sense of a collection of maps; no transition structure
between them is implied, and indeed none exists --- see below. Two
properties of the construction bound what the trajectory can be said to
show. Each \(u^{(\ell)}\) is estimated independently at its own layer
from the same contrastive pairs, with nothing carrying one layer's
direction to the next; the trajectory is therefore a sequence of
independent readings, and ``rotation'' is the difference between
adjacent readings rather than an observed motion. And every cosine
reported below is taken in absolute value, because the sign of a
difference-of-means direction is fixed by the positive/negative
labelling convention and carries no meaning --- so rotation is measured
on the undirected axis, and a 180° reversal registers as no rotation.

\textbf{Terminology.} These terms are used consistently throughout; each
names a distinct object.

\begin{longtable}[]{@{}
  >{\raggedright\arraybackslash}p{(\columnwidth - 2\tabcolsep) * \real{0.3333}}
  >{\raggedright\arraybackslash}p{(\columnwidth - 2\tabcolsep) * \real{0.6667}}@{}}
\toprule\noalign{}
\begin{minipage}[b]{\linewidth}\raggedright
Term
\end{minipage} & \begin{minipage}[b]{\linewidth}\raggedright
Definition
\end{minipage} \\
\midrule\noalign{}
\endhead
\bottomrule\noalign{}
\endlastfoot
\textbf{CAZ segment} & A scored saddle-to-saddle segment of the
separation curve \(S(l)\), around a local peak (§2.2). A locator on
depth, not the concept and not a functional module. \\
\textbf{GEM} & The trajectory record across \textbf{one} CAZ segment:
its layer window, the per-layer directions across it, its settled
direction and its handoff layer. \\
\textbf{atlas} & The set of a concept's GEMs in a model --- one per
segment, mean 2.21 in this corpus. A collection of maps; no transition
structure between them is implied. \\
\textbf{settled direction} \(u_i^{\text{settled}}\) & The trajectory's
terminal reading, taken at the segment's final layer \(\text{end}_i\).
The probe this paper evaluates. \\
\textbf{handoff layer} \(L_{H,i}\) & \(\min(\text{end}_i + 1,\, N-1)\)
--- the first layer outside the segment, derived from the boundary
rather than independently detected (§3.2). For a terminal GEM (45.2\% of
GEMs, §3.2), the segment already ends at \(N-1\), so \(L_{H,i} = N-1\)
is the segment's own last layer, not a layer outside it. \\
\textbf{handoff cosine} &
\(\lvert u_i^{\text{settled}} \cdot u^{(L_{H,i})} \rvert\) --- agreement
across the segment boundary, spanning one layer. \\
\textbf{entry-exit cosine (EEC)} &
\(\lvert u^{(\text{start}_i)} \cdot u^{(\text{end}_i)} \rvert\) ---
total rotation across the segment, reported as a diagnostic. \\
\end{longtable}

\subsubsection{3.2 Handoff Layer
Detection}\label{handoff-layer-detection}

The handoff layer \(L_H\) is operationally defined as:
\[L_H = \min(L_{\text{CAZ\_end}} + 1, \; N - 1)\]

where \(\text{end}_i\) --- the \(L_{\text{CAZ\_end}}\) of the formula,
for region \(i\) --- is the final layer of CAZ region \(i\) as returned
by the scored segmentation (§2.2) --- the saddle point of \(S(l)\)
following the region's separation peak, or the model's final layer in
the case of the deepest region, whose end is set by the model's endpoint
rather than by a detected saddle. \(L_H\) is therefore \textbf{derived
from the region boundary, not detected independently}: no
angular-velocity criterion and no threshold enters its computation.

Earlier versions of this paper described \(L_{\text{CAZ\_end}}\) as the
last layer at which angular velocity exceeded a threshold
\(\varepsilon = 0.05\). \textbf{That criterion is not implemented in the
reference library and does not describe how any result in this paper was
produced}; angular velocity is computed only as a diagnostic (§3.1). The
correction does not change any reported number, because every result was
generated by the segmentation-derived boundary described here.

Conceptually \(L_{H,i}\) is the first layer outside the region. The
settled direction is read at \(\text{end}_i\), the region's last layer;
\(L_{H,i}\) is the first layer at which that direction can be evaluated
against activations it was not estimated on.

The \textbf{handoff cosine} is
\(\left| u_i^{\text{settled}} \cdot u^{(L_{H,i})} \right|\) --- the
agreement between the settled direction, read at \(\text{end}_i\), and
the dominant direction at the first layer outside the region. It is
therefore a single-step continuity measure at the region boundary, not
\textbf{a measure of persistence to the output layer}. It is reported
here per GEM rather than for one selected GEM per pair, giving 1,091
measurements across the 493-pair, 29-model corpus. Across all GEMs the
mean is 0.906 (median 0.947; 73.5\% exceed 0.85).

That figure requires an immediate qualification. The region-detection
procedure assigns every pair exactly one GEM whose segment terminates at
the final layer (§2.2), and for those GEMs \(L_H = N-1\), so the handoff
cosine compares the final layer with itself and is 1.000 by
construction. \textbf{493 of the 1,091 GEMs (45.2\%) are in that
terminal state.} They are not evidence of stability; there is simply no
remaining depth over which the direction could drift. On the 598 GEMs
where \(L_{H,i} < N-1\) and the statistic is therefore informative, the
mean is \textbf{0.828} (median 0.854), with 51.7\% exceeding 0.85 ---
the settled direction survives the step across the region boundary in
roughly half of the cases where that question can be asked at all.
Because the comparison spans a single layer, this paper makes no
measurement of whether the settled direction persists to the output;
that would require comparing \(u_i^{\text{settled}}\) with
\(u^{(N-1)}\), which is not computed here.

Reporting one handoff cosine per concept obscures both facts: it
discards the majority of measured GEMs, and the choice of \emph{which}
GEM to keep determines how often a terminal value is sampled. Within
multi-GEM atlases the handoff cosine also varies systematically with
position within the atlas. Restricting to the informative GEMs, as
everything above does, the mean rises from \textbf{0.814} at the
shallowest GEM to 0.844, 0.863 and 0.905 at successive positions (n =
380, 155, 47 and 12; two further positions carry n = 3 and 1 and are
omitted from this trend for lacking statistical weight, though counted
in the 598 total below), with a mean within-pair spread of
\textbf{0.106}. The gradient is steeper if the terminal GEMs are left in
(0.856 → 0.936 → 0.959 → 0.976, spread 0.192), but that steepness is
largely the construction artifact rather than a property of the
trajectory: the deepest GEM of every atlas is the terminal one, so
including it loads the deep end of the gradient with values that are
1.000 by definition. Either way no single GEM is representative of the
concept's trajectory.

\subsubsection{3.3 Adaptive Ablation
Width}\label{adaptive-ablation-width}

When evaluating probe quality via ablation (§5), we ablate \(w\)
consecutive layers starting at \(L_H\). Ablating more than one layer
risks extending into the final-layer unembedding-preparation zone for
near-final handoffs, diluting the concept signal; the \textbf{near-final
rule} --- when \(L_H/N > 0.85\), set \(w = 1\) --- was introduced to
bound that risk, and §5.2 characterizes its empirical effect on the
391-pair primary sub-corpus, where it is compared against a fixed
\(w = 3\). The stored §5.1 ablations, by contrast, use \textbf{\(w = 1\)
in all 493 cells} at 249--250 pairs per concept --- not the near-final
rule firing (only 109 of the 493 have \(L_H/N > 0.85\) under the
shallowest-GEM convention §5.2 also uses; see §4.3/Table D2 for the
dominant-GEM convention, which reports substantially deeper medians and
is not the same quantity), but simply the operative setting for that
headline result.

An earlier state of the corpus had \texttt{exfiltration} alone at
\(w = 3\) for 28 of 29 models, at 50 pairs rather than 250, and
Llama-3.1-8B on a pre-label-correction extraction. Those files are
superseded and retained under explicit \texttt{superseded\_*} names; all
29 \texttt{exfiltration} ablations were re-run at \(w = 1\) on the
corrected 249-pair pool (§3.5), which is the version every number in
this paper uses.

\subsubsection{3.4 The Settled Direction as
Probe}\label{the-settled-direction-as-probe}

The settled direction \(u_i^{\text{settled}} = u^{(\text{end}_i)}\)
(§3.1) is the GEM-extracted concept probe evaluated at \(L_H\). It is a
unit vector in \(\mathbb{R}^d\) representing the direction in residual
stream space that most discriminates positive from negative pairs at the
layer where that discrimination has stabilized.

To use this probe for downstream tasks --- concept detection, steering,
or testing the Platonic Representation Hypothesis {[}Huh et al., 2024{]}
that models converge on shared representations across architectures
(hereafter PRH alignment) --- it is extracted from the settled-layer
activations and is not retrained on new data. Its validity rests on the
claim that the concept's primary geometric rotation has converged by
\(L_H\) --- it is a stable, single-direction point-estimate of the
concept's encoding at that layer, not a claim that a single static
direction is a complete account of the concept from \(L_H\) onward (see
§9.4).

\subsubsection{3.5 Contrastive Pair
Construction}\label{contrastive-pair-construction}

Pairs are constructed using the multi-model consensus protocol described
in Appendix C. Each positive sentence expresses the target concept in
context; each negative is drawn from a pool of semantically proximate
but concept-absent sentences, with pairs retained only when multiple
independent models agreed on the concept label (the cross-model
consistency filter serves as the ambiguity exclusion criterion). The
construction is designed to isolate the named concept specifically, not
a correlated surface feature; how well that design holds up under
independent human review is reported next.

For the 17 concepts in this paper's corpus (agency, authorization,
causation, certainty, credibility, deception, exfiltration, formality,
moral\_valence, negation, plurality, sarcasm, sentiment, specificity,
temporal\_order, threat\_severity, urgency), pairs were drawn from the
\href{https://github.com/jamesrahenry/Rosetta_Concept_Pairs}{Rosetta
Concept Pairs (RCP) dataset} (DOI:
\href{https://doi.org/10.5281/zenodo.20059650}{10.5281/zenodo.20059650}).
Each concept contributes 250 contrastive pairs; exfiltration contributes
249 following the label correction described below.

\textbf{Pair fidelity.} Human validation of these pairs (companion
validation paper, Henry {[}2026c{]} §2.2/§8.8 and Supplementary §K; 31
raters, 540 analysis-grade ratings over 180 pairs, of which 119 (66\%)
are clean on this first pass) confirms the ``concept-absent'' negative
holds for the large majority of concepts --- 14 of 17 validate at ≥76\%
--- but not uniformly. Three do not: moral\_valence most severely (45\%
of individual ratings valid, yet zero of its ten pairs reach the
≥2-rater clean-consensus threshold --- its negatives are morally
\emph{bad} rather than morally \emph{neutral}), and causation and
deception in a second tier at approximately 63\%. For these the
negatives are opposite-pole passages rather than concept-absent ones, so
an antonym negative still expresses the concept.

This concerns \emph{which} contrast the pairs encode, not whether they
encode one. The pairs separate cleanly either way, and the direction GEM
recovers for these three is a stable, well-estimated axis carrying the
same geometric status as any other in this paper --- it is the pole axis
rather than the concept-presence axis (Henry {[}2026c{]} §8.8, which
reports that low-fidelity pairs still produce a stable, causally-active
CAZ). The handoff-depth and rotation quantities are therefore valid
measurements for all 17 concepts.

\textbf{What the low-fidelity three do not support is the
\emph{semantic} reading of their position in the per-concept orderings
of §4.3, §8.1 and §9.2}: for those concepts the ordering describes where
the pole contrast settles, not where the named concept's presence is
resolved. Automated separability cannot detect this --- a probe
separates antonym pairs perfectly --- so the fidelity map rests on human
rating (Henry {[}2026c{]} §8.8). The aggregate results are unaffected,
and we verify rather than assert this: recomputing the headline
precision result (§5.1) over only the 14 high-fidelity concepts gives
281/406 (69.2\%), against 341/493 (69.2\%) over all 17 --- a difference
of 0.0 percentage points.

\textbf{Exfiltration label correction.} The exfiltration pairs in the
original corpus carried inverted positive/negative labels for
approximately 72\% of pairs. The concept was re-extracted after the
correction and contributes 249 pairs against 250 for the other sixteen
concepts. This is a corpus-level fix shared with the companion
validation paper (Henry {[}2026c{]} §D); results reported here use the
corrected extraction.

\textbf{Generator robustness.} The companion validation paper tests
directly whether any single generator is load-bearing for the recovered
geometry: leave-one-generator-out analyses across model pairs and all 17
concepts (Henry {[}2026c{]}) leave the results intact. We do not repeat
that analysis here and rely on it.

\textbf{Corpus provenance.} All 14 generators are themselves transformer
language models, so the corpus could in principle reflect convergence on
how such models render a contrastive distinction rather than on the
concept itself. We disclose this as a scope condition rather than
resolving it. The direct test --- refitting on a human-authored
contrastive set paired with a register-matched control --- is specified
here and not run; no result in this paper is conditioned on its outcome.

\begin{center}\rule{0.5\linewidth}{0.5pt}\end{center}

\subsection{4. Rotation and Settling}\label{rotation-and-settling}

This section characterizes concept-direction rotation and settling
empirically across our 29-model, 17-concept corpus before turning to
validation. The handoff layer, used throughout as the post-settling
extraction point, is characterized here only as the point where our data
shows rotation has stopped --- not established as privileged for
reading, extraction, or anything else within that process --- see §5.5.

\subsubsection{4.1 Rotation Is Ubiquitous}\label{rotation-is-ubiquitous}

A concept's direction in the residual stream is not a fixed point
waiting to be read off at some depth. It has two properties this paper
treats as inseparable: a \emph{depth} at which it forms (its CAZ) and a
\emph{trajectory} of rotation while it forms. This section establishes
the second half of that claim empirically: across every architecture and
concept type in this corpus, the direction is not stable while a concept
is being assembled --- it rotates substantially, and only stops once
assembly is complete.

Table 1 shows the mean and median entry-exit cosine (EEC) --- the cosine
similarity between the concept direction at CAZ entry and at CAZ exit
--- across all 493 concept × model pairs (29 models × 17 concepts).

\textbf{The estimator's own noise floor.} A rotation statistic is only
interpretable against the agreement two \emph{independent} estimates of
the same direction achieve, so we measured it: each concept's pairs are
split in half, the difference-of-means direction is estimated separately
on each half at the same layer, and the two are compared. Across seven
models spanning six families (pythia-410m, pythia-2.8b, gpt2-large,
opt-1.3b, Qwen2.5-3B, Llama-3.2-3B, phi-2) the peak-layer floor is
\textbf{0.966--0.975, mean 0.970} --- near-constant across architecture
and scale, so a single value serves as the denominator rather than a
per-model table. If a concept did not rotate at all, its EEC would
measure this floor; the observed mean of 0.285 is far below it, so the
rotation reported here is a property of the representation and not of
the estimator.

This comparison is asymmetric: the 0.970 floor is measured at the peak
layer, where SNR is highest, while EEC compares an entry-layer direction
--- noisier by construction, before CAZ onset (§2.3) --- against an
exit-layer one. An entry-layer floor, the one that actually matches what
EEC compares, is not measured here; if it runs lower than 0.970, as a
noisier estimate would predict, the true no-rotation baseline for this
specific contrast is lower too, and 0.285 is not as far below it as the
0.285-vs-0.970 framing suggests --- though a gap this large is unlikely
to close entirely.

The one exception is gemma-2-2b, whose floor is \textbf{0.760} --- the
only Gemma-2 model this split-half check was run on; gemma-2-9b's floor
is not measured here --- see §7.4.

\textbf{Table 1: Entry-exit cosine statistics (all 493 pairs)}

\begin{longtable}[]{@{}ll@{}}
\toprule\noalign{}
Statistic & Value \\
\midrule\noalign{}
\endhead
\bottomrule\noalign{}
\endlastfoot
Mean EEC & 0.285 \\
Median EEC & 0.292 \\
\% pairs with EEC \textless{} 0.5 & 91.5\% \\
\% pairs with EEC \textless{} 0.1 & 15.4\% \\
Mean max rotation per layer & 0.380 \\
Mean per-GEM rotation per layer & 0.295 \\
\end{longtable}

The low EEC values confirm the core premise: across architectures and
concept types, concept directions rotate substantially during assembly.
The 15.4\% of pairs with near-orthogonal rotation (EEC \textless{} 0.1)
represent cases where the pre-CAZ and post-CAZ directions share almost
no orientation --- probing at CAZ entry would extract a direction
orthogonal to the settled post-assembly direction.

\textbf{Aggregation and its limits.} A concept × model pair has a mean
of 2.21 GEMs per pair (distribution: 1 GEM in 113 pairs, 2 in 225, 3 in
108, 4 in 35, 5 in 9, 6 in 2, 7 in 1), and every GEM has its own
entry-exit cosine. All EEC figures in this paper are computed by
averaging across a pair's GEMs and then across the 493 pairs, so each
concept × model pair contributes equally regardless of how many GEMs it
has; pooling GEMs directly would let finely-partitioned pairs dominate
(that alternative gives 0.342). One caveat must be stated plainly:
\textbf{EEC is sensitive to how finely the detector partitions a
concept.} A GEM's EEC correlates with the number of layers it spans at
\emph{r} = −0.53 --- narrow regions have less room to rotate and
therefore score higher --- so mean EEC across a corpus is partly a
statement about segmentation granularity, and would move under different
region-detection settings (§2.2) with no change to any model. The
per-layer rotation rate is the more segmentation-robust quantity: the
maximum single-layer rotation is near-independent of GEM span (\emph{r}
= −0.07), and it is reported alongside EEC in Table 1 for that reason.

\subsubsection{4.2 Rotation Across
Architectures}\label{rotation-across-architectures}

Rotation is present in every model in the corpus; the per-model figures
are in \textbf{Appendix D, Table D1}. No architecture is immune: gpt2
and gpt2-medium show the lowest mean EEC (0.103, 0.104), rotating almost
completely across their CAZs. The model with the highest mean EEC
(Qwen2.5-3B: 0.489) still has EEC below 0.5. Rotation rate tracks depth
as expected --- the shallowest models rotate fastest per layer (gpt2
0.586, pythia-70m 0.541) because they must complete assembly in fewer
layers, while the deepest rotate slowest (gemma-2-9b 0.118).

\subsubsection{4.3 Handoff Depth by
Concept}\label{handoff-depth-by-concept}

Per-concept mean handoff relative depth is tabulated in \textbf{Appendix
D, Table D2}. It reports \(L_H / N\) per concept type, across the 23
models of the primary sub-corpus (Appendix A), where \(L_H\) is the
handoff layer of the dominant CAZ region --- the highest-scoring GEM in
each concept's atlas, which need not be the deepest GEM in a multi-GEM
atlas (a distinct comparison from §5.2's shallowest-vs-dominant overlap
statistic, 34.4\%, which this figure should not be read against).

These figures --- and the §5.4 GEM-count distribution --- are
regenerated from the frozen extraction store by
\texttt{gem/regenerate\_table3\_handoff\_depth.py} in the analysis
release. With N=250 pairs, the handoff layer distribution shifts
substantially later relative to pilot data: most concepts settle in the
final 15--30\% of model depth, with specificity, plurality, and
sentiment showing the earliest handoffs.

\begin{itemize}
\tightlist
\item
  \textbf{Shallow (\textless0.72)}: specificity, plurality, sentiment
\item
  \textbf{Mid (0.72--0.86)}: moral\_valence, formality, credibility,
  deception, sarcasm, urgency, agency, negation, exfiltration, causation
\item
  \textbf{Deep (≥0.86)}: authorization, temporal\_order, certainty,
  threat\_severity
\end{itemize}

This banding is computed on the \textbf{mean} column of Table D2. The
median column (also given there) diverges for several concepts --- most
sharply for moral\_valence, whose median (0.917) sits in the Deep range
despite a Mid-range mean --- because per-model depth is not
symmetrically distributed; the discussion of specific concepts below
switches to median where noted.

\emph{Pair fidelity (§3.5): moral\_valence, causation and deception are
the three concepts whose negatives are opposite-pole rather than
concept-absent. Their depths are valid measurements of the axis the
pairs induce, but that axis is the pole contrast --- read their position
in this ordering accordingly.}

The depth ordering differs from pilot data in an important way: with
N=250 pairs the separation curve \(S(l)\) is estimated more stably and
its saddle points move later, so region ends --- and with them the
handoff layers --- sit further toward the final layer for most concepts
in most models (see §9.3). Part of this pattern is structural rather
than empirical: the deepest region's end is set to the model's final
layer by construction (§2.2), so every concept contributes one handoff
at or adjacent to \(N-1\) regardless of its separation curve. The high
median values (≥ 0.917 for 12 of 17 concepts) reflect that more than
half of models have their handoff in the final 8--10\% of depth for
those concepts.

By median rather than mean, the shallowest-settling concept is
specificity (median 0.607), with plurality, formality, and credibility
next (median 0.750) and sentiment following (0.833); these are the
exceptions to the otherwise-late pattern. Among the deepest concepts,
threat\_severity (mean 0.911) and certainty (0.895) sit deepest; the
difference from mid-band concepts is smaller than in pilot data.

This ordering sits in the same region as the moderate cross-architecture
depth-ordering tendency the companion validation paper reports (median
\(\tau = 0.404\); {[}Henry, 2026c §3{]}). The two orderings should not
be equated: that paper ranks concepts by CAZ dominant-peak depth and
this one by GEM handoff depth, and they disagree on individual concepts
--- exfiltration is that paper's deepest concept but ranks 12th of 17
here, while certainty is mid-pack there and second-deepest here.

\begin{center}\rule{0.5\linewidth}{0.5pt}\end{center}

\subsection{5. Validation}\label{validation}

\textbf{Corpus design.} §5.1's headline comparison and the §5.3/§5.5
controls run on the \textbf{29-model, 493-pair handoff corpus}; §5.2's
adaptive-width experiment was calibrated on the nested \textbf{23-model,
391-pair primary sub-corpus} (Appendix A) and is not re-run on the six
additions; §5.6's random-window null runs on \textbf{thirteen models} at
the same 250 pairs per concept, in two batches (30 windows for six
models, 50 for the other seven) because of the added ablation cost. Each
section states the corpus it reports.

\subsubsection{5.1 Handoff Layer vs.~Peak Layer: Ablation
Experiment}\label{handoff-layer-vs.-peak-layer-ablation-experiment}

\textbf{Setup.} For each concept × model pair, we ablate two arms with
matched site count, one site per CAZ region in the concept's atlas
(§3.1). The \emph{handoff} arm ablates the settled direction
\(u_i^{\text{settled}}\) at each region's handoff layer \(L_{H,i}\); the
\emph{peak} arm ablates the region-local direction at each region's own
CAZ peak layer --- a per-region peak, not a single corpus- or model-wide
peak. An \(n\)-region atlas therefore ablates \(n\) sites in both arms
simultaneously (mean 2.21 regions per pair, §4.1); single-GEM atlases
(one region) ablate one site per arm, the degenerate case in which both
arms coincide with the earlier single-layer description. We then measure
the fractional reduction in concept separation (the \emph{retained
percentage} after ablation, lower = better ablation = stronger probe).

The ablation operator is directional ablation as introduced by Arditi et
al.~{[}2024{]}: the concept direction is removed from all residual
stream activations at the probe layer, so that for activation
\(h \in \mathbb{R}^d\) the ablated activation is
\(h' = h - (h \cdot u)\,u\), which zeroes the component along \(u\)
while preserving all orthogonal components. We apply it simultaneously
at each arm's target layers as a probe-quality measurement, where Arditi
et al.~apply it at every layer as a behavioral intervention.

Formally, for each arm's set of target layers, the corresponding
orthogonal projection \(P_u\) is applied to all residual stream vectors
at each target layer, and separation is then measured at every region's
CAZ peak and at the model's \textbf{final layer}; we report the
final-layer figure throughout. A \textbf{handoff improvement} is
recorded when the handoff arm achieves lower final-layer retained
percentage (greater separation suppression) than the peak arm. Both arms
are therefore scored against the same final-layer baseline, so retained
percentages are directly comparable.

\textbf{Depth confound}: \(L_H\) is always ≥ \(L_{\text{peak}}\) (the
handoff layer is post-CAZ; the peak is inside the CAZ). The handoff
probe improvement rate could partly reflect a depth-of-extraction
advantage --- later layers carry more processed information --- rather
than directional settling per se. Because the measurement is taken at
the output, the confound is if anything sharper than a within-layer
comparison would make it: ablating later leaves fewer remaining layers
in which the model can reconstitute the concept, so the deeper
intervention is mechanically advantaged.

The direction-specificity control (§5.3) does \textbf{not} address this.
It establishes that the suppression is carried by the concept direction
rather than by an arbitrary direction --- which is not in dispute ---
but both arms of the handoff-versus-peak comparison use concept
directions, so a random-direction null cannot separate depth from
settling. The control that does speak to location randomizes
\emph{where} the ablation window sits --- a width-matched random-window
null, reported in §5.6 on thirteen models. It does not hold the
direction fixed: the concept direction is re-estimated at each random
window, so the comparison is between GEM's layer read with GEM's
direction and an arbitrary layer read with the direction available
there. That is the question worth asking of a layer-selection method,
but it means the null bounds location and estimate quality jointly
rather than location alone.

A control comparing the atlas's handoff layers against layers chosen
without the GEM settling criterion, at matched depth and matched site
count, speaks to the settling contribution more directly. That
experiment is reported in §5.5, and it \textbf{largely removes the
effect}: the handoff layers win in 263/493 pairs (53.3\%, per-model
win-rate Wilcoxon \(p = 0.456\)) corpus-wide; one of the eight
roster/aggregator combinations tried on restricted rosters clears
\(p<0.05\) (this paper's own roster and aggregator, \(p=0.037\)), but
not the other seven, which is why we do not treat any residual as
established rather than directional (§5.5).

\textbf{Results.} The headline result is now reported over the expanded
29-base-model corpus (§5.5's coverage --- the 23-model primary
sub-corpus plus six additional base models; four further large models
could not be included for VRAM reasons and one mixture-of-experts model
is excluded by scope, see Appendix A and §5.5): across 493 concept ×
model pairs (29 models × 17 concepts), the handoff-layer GEM probe is
more precise than the peak-layer probe in 341/493 cases (69.2\%).
Because the two Gemma-2 models are a documented measurement limit rather
than a negative result (§7.4), we report the headline both ways:
excluding them gives 326/459 (71.0\%) across the remaining 27 models.
The Gemma-2 rows are retained throughout, following the companion
paper's practice {[}Henry, 2026c §6.9{]}; excluding them would raise the
reported rate, not lower it.

A Wilcoxon signed-rank test on per-model handoff-preference proportions
(N=29 models) gives W=356, p=1.34×10⁻³ (one-sided) --- significant at
the model level. The within-model correlation structure is evident from
per-model improvement rates (Llama-3.1-8B: 17/17 = 100\%; pythia-1.4b:
17/17 = 100\%; gpt2-large: 17/17 = 100\%; gpt2-xl: 17/17 = 100\%;
pythia-12b: 16/17 = 94\%); trials are not exchangeable within a model,
and treating them as independent Bernoulli draws would overstate the
evidence; see §9.3.

The companion validation paper reports 304/442 (68.8\%) for the same
comparison {[}Henry, 2026c §6.5{]}. That figure is \textbf{not}
independent corroboration of the rate here: both papers read the
\emph{same stored} field (\texttt{comparison.handoff\_better}) from the
same ablation run, and the companion's 442 cells are a strict subset of
this corpus --- its 26 models are 26 of these 29. The two rates agreeing
to 0.4 pp is arithmetic, not replication. It is reported here only to
show that the same computation restricted to the companion's roster does
not change the picture; no independent test of §5.1 exists anywhere in
this program.

\textbf{There is no MHA/GQA cohort split at this corpus size.} 13 of 17
MHA models and 6 of 9 GQA models prefer handoff on a strict majority of
concepts --- statistically indistinguishable rates (Fisher's exact,
one-sided p = 0.462). phi-2 and the two Gemma-2 models are excluded from
this comparison; Appendix A classes phi-2 as ``Other'' for its parallel
attention block and Gemma-2 as alternating attention. The
parallel-attention exclusion is applied as Appendix A's family
classification, not as a mechanical rule: the Pythia family also
interleaves attention and MLP in parallel, and excluding all eight
Pythia models on that basis would remove the MHA cohort's majority. We
flag the tension rather than resolve it, and note that the cohort
comparison is not load-bearing for any claim in this paper --- it is
reported to retract a previous one.

The published 23-model-corpus cohort split (11/13 MHA vs.~2/7 GQA,
p=0.022) does not survive at this corpus size --- both cohorts now
mostly prefer handoff; see §9.3 for why (re-measurement instability in
three models, not simply corpus growth).

GEM-handoff is preferred on a strict majority of concepts in \textbf{21
of the 29 models}. The 8 below-majority models --- pythia-70m,
pythia-160m, gpt2, gpt2-medium, Qwen2.5-0.5B, Qwen2.5-3B, Qwen2.5-14B,
gemma-2-2b --- span both attention cohorts, confirming this is not a
cohort effect. gpt2 is the documented structured failure for
independently identifiable reasons (§7); the other three below-majority
MHA models (pythia-70m, pythia-160m and gpt2-medium) are small-scale
cases, not evidence of an MHA-wide pattern --- 13 of the 17 MHA models
clear a strict majority. The original 23-model primary corpus (Appendix
A) remains the basis for §5.2's adaptive-width experiment below, which
was calibrated against it and is not re-run here.

Improvement rates by scale across the full 493-pair corpus:

\begin{itemize}
\tightlist
\item
  Models \textless{} 500M: 47\% (56/119) {[}7 models: pythia-70m, gpt2,
  opt-125m, pythia-160m, opt-350m, gpt2-medium, pythia-410m{]}
\item
  Models 500M--3B: 80\% (163/204) {[}12 models{]}
\item
  Models \textgreater{} 3B: 72\% (122/170) {[}10 models{]}
\end{itemize}

These are not a floor followed by a plateau. The sub-500M band is
\textbf{bimodal}: gpt2 (6\%), gpt2-medium (6\%), pythia-160m (29\%) and
pythia-70m (41\%) against opt-125m (82\%), opt-350m (76\%) and
pythia-410m (88\%) --- three of seven at or above the \textgreater3B
rate of 72\%. The middle band is the highest of the three, so the trend
is not monotone, and across all 29 models the correlation between scale
and handoff preference is weak (Pearson r = +0.34 on log parameters). We
therefore make no scale claim.

\textbf{gpt2 ablation failure.} gpt2 is a second structured failure,
distinct from opt-6.7b (§7). In 12/17 pairs, gpt2's handoff ablation
produces retained\_pct \textgreater{} 100\% --- meaning ablation at the
handoff layer pathologically \emph{increases} concept separation rather
than suppressing it. gpt2-medium shows the same pathology in 10/17 pairs
(59\%), and no other model in the corpus has a single such pair, so the
GPT-2 family contributes both of the corpus's lowest scores. Excluding
both from the scale breakdown lifts the sub-500M band from 47\% (56/119)
to \textbf{64\% (54/85)}, close to the \textgreater3B rate. What remains
below the corpus mean is the two smallest Pythias (pythia-70m 41\%,
pythia-160m 29\%), and depth does not account for them: at 12 layers
this corpus contains gpt2 (6\%), pythia-160m (29\%) and opt-125m (82\%),
and at 24 layers gpt2-medium (6\%) and pythia-410m (88\%). Matched on
depth, the spread is nearly the full range. We report the grouping as an
observation about which models fail --- a documented ablation pathology
in one family, plus the two smallest Pythias --- and not as a tested
mechanism. pythia-410m (88\%) exceeds the corpus mean and is the
exception in this bucket. Results including gpt2 are reported for corpus
completeness.

opt-6.7b achieves 13/17 improvements (76\%) at N=250 --- above the
corpus mean and not a failure mode (§7). The primary statistics include
both OPT and gpt2.

The magnitude of improvement, on the original 391-pair primary
sub-corpus (not recomputed at 493 pairs): for pairs where the handoff
identifies a better probe location, the mean retained percentage
improvement is 20.4 percentage points (handoff: mean 32.8\% retained
vs.~peak: mean 53.2\% retained for improved cases). For cases where the
peak is the better location (132 pairs), the mean degradation is 16.9pp
(handoff: mean 61.8\% retained vs.~peak: mean 44.9\% retained for those
cases). The net expected improvement per pair (improvement × rate −
degradation × non-improvement rate) = 20.4 × 0.662 − 16.9 × 0.338 =
+7.8pp, consistent with the observed overall mean difference (+7.78pp
from the aggregate).

\subsubsection{5.2 Adaptive Width
Experiment}\label{adaptive-width-experiment}

\textbf{Setup.} The adaptive width experiment tests whether the
near-final rule (w=1 when \(L_H/N\) \textgreater{} 0.85) improves probe
quality over fixed w=3. We run the ablation across 391 pairs (23 models
× 17 concepts, the primary corpus) and compare adaptive vs.~fixed
reduction percentages. One scope note on the trigger: \(L_H/N\) here is
evaluated on the \emph{first} GEM in each concept's atlas --- the
shallowest segment --- because the generating script indexes
\texttt{ablation\_targets{[}0{]}}, and that list is every GEM in index
order rather than a ranked selection. The headline comparison of §5.1,
by contrast, ablates every GEM's handoff layer. Across this corpus the
shallowest GEM is also the highest-scoring one in 34.4\% of multi-GEM
atlases, so for most concepts the trigger is evaluated on a different
segment than the intervention it modifies. We report the rule as
calibrated rather than re-deriving it, and flag the mismatch rather than
leaving it implicit; the results below are scoped to this corpus and are
not used elsewhere in the paper.

\textbf{Results.} Table 2 summarizes the findings.

\textbf{Table 2: Adaptive width experiment results (391 pairs, 23 models
× 17 concepts)}

\begin{longtable}[]{@{}
  >{\raggedright\arraybackslash}p{(\columnwidth - 4\tabcolsep) * \real{0.3137}}
  >{\raggedright\arraybackslash}p{(\columnwidth - 4\tabcolsep) * \real{0.1765}}
  >{\raggedright\arraybackslash}p{(\columnwidth - 4\tabcolsep) * \real{0.5098}}@{}}
\toprule\noalign{}
\begin{minipage}[b]{\linewidth}\raggedright
Rule triggered
\end{minipage} & \begin{minipage}[b]{\linewidth}\raggedright
N pairs
\end{minipage} & \begin{minipage}[b]{\linewidth}\raggedright
Mean delta vs.~fixed w=3
\end{minipage} \\
\midrule\noalign{}
\endhead
\bottomrule\noalign{}
\endlastfoot
Near-final (\(L_H/N\) \textgreater{} 0.85) & 79 triggered (60 improved)
& \textbf{+7.44pp (improved cases only)} \\
--- net effect on all 79 triggered cases & 79 & \textbf{≈+4.60pp}
(algebraically implied by the two rows below, since the default row is
exactly 0.00pp: 391 × 0.93 ÷ 79) \\
Default (w=3, unchanged) & 312 & +0.00pp \\
Overall & 391 & +0.93pp \\
\end{longtable}

The near-final rule triggers in 79/391 cases (20.2\%), distributed
across multiple models with late-layer assembly patterns. The highest
trigger counts are gpt2 (11 concepts), pythia-12b (7), Mistral-7B-v0.3
(7), pythia-1b (6), phi-2 (6), and pythia-1.4b (6). The pattern reflects
architectures where the primary handoff occurs in the final 15\% of
model depth.

When triggered, the near-final rule improves probe quality in
\textbf{60/79 cases (75.9\%)} with a mean improvement of +7.44pp for
cases showing improvement --- but that figure excludes the 19 failures
(to −27.9pp); the net effect across all 79 triggered cases, the number a
rule should be headlined by, is \textbf{≈+4.60pp} (Table 2). The
underlying mechanism for improvements: at \(L_H/N\) \textgreater{} 0.85,
averaging over 3 layers captures the handoff layer plus 1--2
post-handoff layers where the representation begins to be absorbed into
the unembedding computation, diluting the concept signal. Width 1
extracts the direction exactly at the convergence point.

\textbf{gpt2 failure.} The near-final rule fails systematically for gpt2
(12 layers): 10 of 11 triggered cases do not improve --- 3 show severe
degradation (−12.5, −17.6, −27.9pp for threat\_severity, formality, and
temporal\_order respectively) and 7 show zero delta. One case (negation,
+9.5pp) improves. In gpt2, near-final handoffs at \(L_H/N\) ≈ 0.92
correspond to layer 11 --- leaving only one remaining layer before the
unembedding projection. Width 1 at this depth extracts a direction
heavily contaminated by unembedding-preparation computations. A
depth-corrected near-final rule (\(L_H/N\) \textgreater{} 0.85 AND N ≥
20) excludes gpt2 (N=12, 11 triggered cases) and pythia-1b (N=16, 6
triggered cases) --- 17 total excluded cases. The remaining 62 triggered
cases show 53/62 improvements (85.5\%) with mean delta +6.80pp for cases
showing improvement. This refined rule is recommended for future use;
the results above report the simple threshold to document the failure
mode.

Excluding gpt2 from the triggered cases, 8 remaining degradation cases
exceed −0.5pp: pythia-410m/sentiment (−12.1pp), phi-2/urgency (−6.5pp),
phi-2/causation (−0.9pp), phi-2/plurality (−0.7pp), and
Mistral-7B/exfiltration, temporal\_order, plurality, causation (−0.7 to
−1.5pp). These likely reflect concepts with noisy handoff detection
rather than rule failure.

Behavioral ablation at scale (the behavioral ablation script) is
designed but not run; the current results measure separation suppression
at the model's final layer rather than downstream output change.

\subsubsection{5.3 Direction-Specificity
Control}\label{direction-specificity-control}

\textbf{Motivation.} Ablation experiments show that projecting out the
GEM-extracted direction suppresses concept separation. A sceptical
reviewer can ask: is this because the \emph{direction} carries
concept-specific information, or simply because projecting out
\emph{any} direction from the residual stream causes incidental
separation loss? This control answers that directly by comparing
concept-direction ablation against ablation in 10 independently sampled
random directions per concept × model pair.

\textbf{Setup.} For each pair where the concept-direction ablation
produces non-zero separation reduction, we sample 10 random unit vectors
uniformly from the residual stream's \(d\)-dimensional sphere, project
each out independently at the handoff layer, and measure the resulting
separation reduction. The concept-direction reduction and the
distribution of 10 random reductions are compared. At C=17, analysis
covers 472 concept × model pairs across all 29 base models.

Pairs with concept-direction reduction ≤ 0 are excluded as uninformative
for this comparison, dropping most of gpt2's and gpt2-medium's concepts
--- the documented structured-failure case (§7) where handoff-layer
ablation frequently increases rather than decreases concept separation.
The 21 excluded cells are one short of the 22 pathological cells §5.1
reports for the two models combined (12/17 for gpt2, 10/17 for
gpt2-medium); we have not traced the single-cell discrepancy, so we do
not assert that every one of the other 27 models has full 17-concept
coverage here, only that the exclusions are concentrated in the GPT-2
family.

The OPT family initially failed to load under this stack's default
torch/transformers version (CVE-2025-32434's legacy-checkpoint
mitigation) and was recovered by pinning torch 2.6.0 in the extraction
flow.

\textbf{Results.} Across 472 pairs (C=17, 29 models):

\textbf{Table 3: Direction-specificity control summary (472 pairs)}

\begin{longtable}[]{@{}ll@{}}
\toprule\noalign{}
Metric & Value \\
\midrule\noalign{}
\endhead
\bottomrule\noalign{}
\endlastfoot
Mean concept-direction reduction & \textbf{66.3\%} \\
Mean random-direction reduction & \textbf{0.07\%} \\
Median specificity ratio & \textbf{2,552.4×} \\
\end{longtable}

The concept direction suppresses separation at 2,552.4× the median rate
of a random direction. \textbf{That ratio is not a calibrated effect
size, and we do not read it as one.} A random unit vector in \(d\)
dimensions has expected squared overlap \(1/d\) with any fixed
direction, so the random-ablation baseline measures dimensionality
rather than concept structure --- and it behaves that way here. Across
the 29 models the mean random reduction scales as \(d^{-0.805}\)
(Pearson \(r = +0.562\) against \(1/d\)), every model's value falls
within a factor of four of its own \(1/d\), and the corpus mean of
0.07\% sits inside the 0.020--0.195\% band that \(1/d\) spans for hidden
widths of 512--5120.

The specificity ratio is consequently approximately (concept reduction ×
\(d\)): regressing the per-pair ratio on hidden dimension gives a
log-log slope of \textbf{+1.015}, so doubling a model's width doubles
its reported ratio for arithmetic reasons alone. We therefore report the
ratio as a descriptive figure only.

Broken down by architecture cohort, every cohort beats all ten of its
random seeds in 100\% of pairs, and mean concept-direction reduction
ranges from 38.6\% (Gemma) to 71.0\% (phi-2); the per-cohort table is in
\textbf{Appendix D, Table D3}.

Phi-2 is reported as its own cohort rather than folded into MHA,
consistent with §5.1's Appendix A convention of excluding it from the
MHA/GQA split as a parallel-attention architecture. (Eight of the 472
pairs --- 1.7\% --- have a formally infinite z-score because their
random-direction ablations produced zero measurable variance across all
10 seeds; none falls at a cohort's median position, so the reported
medians above are unaffected.)

The ratio column above should not be read across cohorts: cohorts differ
in mean hidden width, and the ratio tracks width at a log-log slope of
+1.015. The concept and random columns, and the \textgreater all-seeds
column, are the interpretable ones.

Gemma has the \emph{lowest} median specificity ratio of the four cohorts
(1,249.9× vs.~1,451.8--3,175.9× elsewhere), alongside its lowest
absolute reduction (38.6\%). Neither this pattern nor the original
7-concept pilot's opposite one --- Gemma with the \emph{highest}
specificity ratio (863×), read at the time as more geometrically precise
concept encoding --- should be read as a fact about Gemma-2's encoding.
Both are architectural readings of a direction estimate that is
under-resourced at N=250 for this family: the single-direction ablation
these figures are computed from removes essentially no decodable
information on gemma-2-2b, so its magnitude and specificity are
uninformative about the underlying representation. See §7.4.

\textbf{Empirical p note.} With 10 random seeds, the minimum achievable
empirical p for ``concept beats all seeds'' is \(1/11 \approx 0.09\),
which does not reach \(p < 0.05\) per pair. The substantive conclusion
is clear independent of formal p-value, and it rests on the seed
comparison rather than on the ratio: \textbf{all four cohorts beat every
one of their 10 random seeds on 100\% of pairs.} That statistic is
scale-free --- unlike the ratio, it does not vary with hidden width ---
and it is what refutes the reading this control was built to test,
namely that projecting out \emph{any} direction at the handoff layer
would destroy separation. It does not; random directions reduce
separation at the \(1/d\) chance level while the concept direction
removes a mean of 66.3\%.

The per-cohort distributions are shown in Figure 2; the random-direction
distributions collapse to near zero, which is why the panel renders as
box-and-whisker rather than as resolvable densities.

\begin{figure}
\centering
\includegraphics{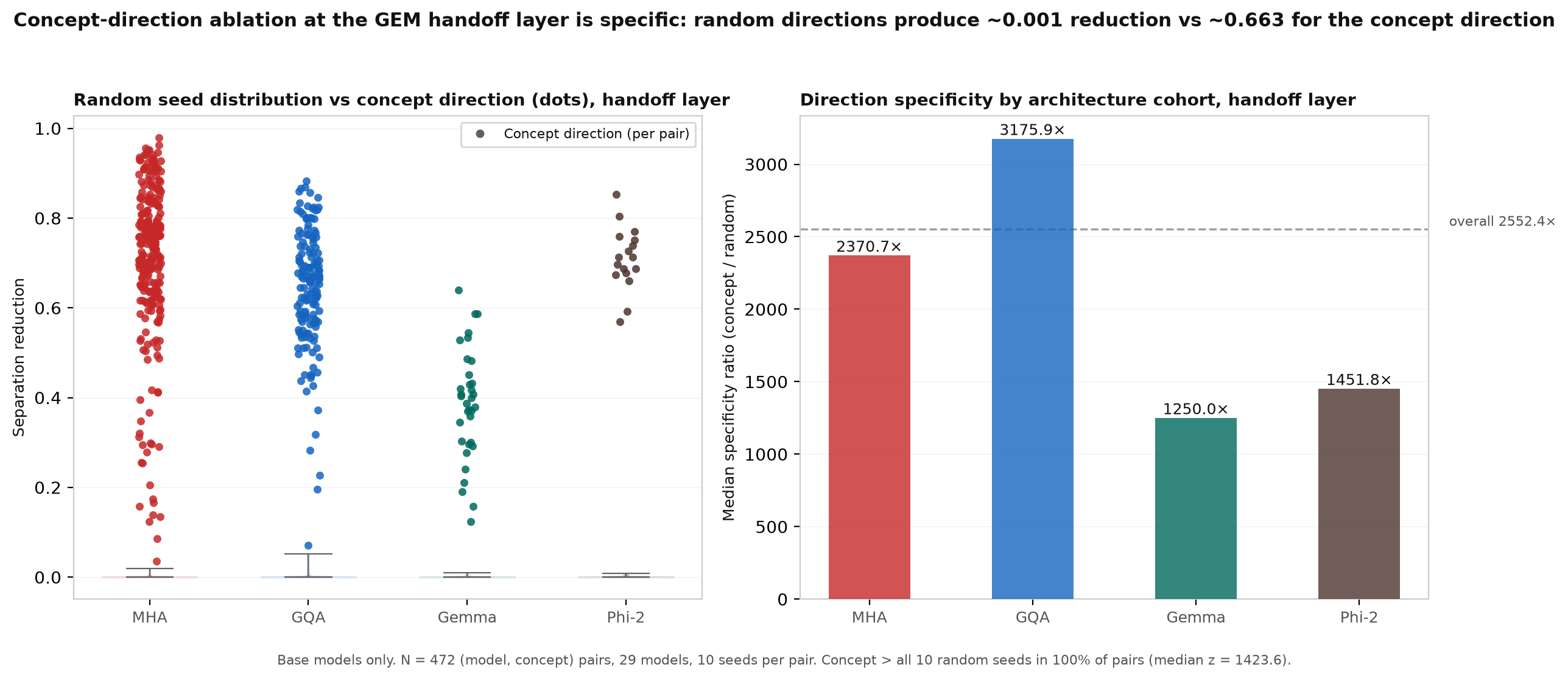}
\caption{Figure 2: Box-and-whisker summary of random-direction
separation reduction (10 seeds per pair) by architecture cohort, with
concept-direction reduction overlaid as individual points, at the
handoff layer, C=17 (472 pairs, 29 models, mean 66.3\%/median 2,552.4×,
cohorts MHA/GQA/Gemma/Phi-2).}
\end{figure}

\subsubsection{5.4 Ablating Multiple GEMs}\label{ablating-multiple-gems}

\textbf{Motivation.} The preceding validation treats each concept's
atlas as a single GEM --- one region's handoff layer, derived from that
region's boundary (§3.2). In practice, most concepts have several GEMs:
the companion validation paper {[}Henry, 2026c{]} finds a mean of 2.20
CAZ regions per concept × model pair, and this paper's own segmentation
gives 2.21 GEMs per pair (§4.1); each region can produce a detectable
handoff. A multi-GEM atlas raises a structural question: are the GEMs
independent encodings of the same concept (redundant), or is each GEM a
precondition for the next? The answer determines whether GEM is
documenting a distributed relay or selecting among redundant
alternatives.

\textbf{Setup.} GEM-count distribution, provenance and the
cross-architecture comparison are in \textbf{Appendix D.4}. We ran a
full permutation ablation on three models --- pythia-6.9b (32 layers),
gpt2-xl (48) and Qwen2.5-7B (28) --- at the corpus standard of
\textbf{250 pairs per concept}: for every concept with multiple GEMs, we
tested every non-empty subset of GEMs (\(2^N - 1\) subsets per concept),
applying directional ablation simultaneously at all GEMs in the subset
and measuring separation reduction at every GEM's handoff layer and at
the final layer. That yields, per concept: which single GEM drives
final-layer separation most; whether ablating all GEMs together exceeds
the best single one (synergy); and whether ablating only the shallower
GEMs reduces separation at the deepest GEM's handoff layer
(cross-disruption). Width = 1 at each handoff layer. 42 multi-GEM
atlases across the three models.

\textbf{Results.}

\emph{The deepest GEM is the most effective single site (40/42).} In 40
of 42 multi-GEM atlases the single ablation producing the greatest
final-layer reduction targets the deepest GEM --- 15/15 on pythia-6.9b,
11/11 on gpt2-xl, 14/16 on Qwen2.5-7B. This is exactly what §5.1's depth
confound predicts on its own: the deepest GEM's handoff sits at or
adjacent to \(N-1\) by construction, separation is measured at the final
layer, and ablating closer to that measurement point leaves the model
fewer remaining layers to reconstitute the concept. We do not have a
depth-matched control for this specific comparison (unlike §5.1's, which
does; §5.5), so 40/42 should be read as consistent with a depth account,
not as evidence against one.

\emph{Cross-disruption is present in every model (37/42).} Ablating only
the shallower GEMs reduces separation at the deepest GEM's own handoff
layer in 11/15 (mean 0.253), 10/11 (0.15) and 16/16 (0.40) atlases
respectively. Ablating shallower GEMs therefore changes what is
measurable at the deepest one; we report the association and do not
infer a causal ordering from it.

\emph{Synergy is model-dependent, and this is the result that most
needed the larger sample.} On the two MHA models, ablating all GEMs
together suppresses \textbf{less} than the best single GEM --- mean
synergy −0.103 on pythia-6.9b (\textbf{0 of 15} atlases positive) and
−0.089 on gpt2-xl (\textbf{0 of 11}). On Qwen2.5-7B the sign
\textbf{inverts}: mean +0.061, with \textbf{13 of 16} atlases positive.
There, ablating more sites does suppress more.

We flag two limits on reading that split. It is three models, only one
of which is GQA, and this paper has already retracted one
architecture-cohort claim on a larger sample (§5.1) --- so we do not
offer it as a cohort effect. And the negative-synergy measurement cannot
distinguish genuine anti-synergy from patch interference: ablating an
upstream GEM shifts the representation that a downstream GEM's stored
direction was estimated on, so a stale direction under-ablates.
Resolving that needs the adaptive-direction protocol (§9.4).

\emph{Single-GEM ceiling.} Even at the deepest and most effective GEM,
single-GEM ablation does not approach complete suppression on any model,
consistent with §5.1 (mean retained 32.8\% for improved cases).

\textbf{What we take from this, and what we do not.} Ablation does not
cleanly remove the concept: at the best single site, on every model
tested, suppression is partial. Whether intervening at \emph{more} sites
helps depends on the model --- it does on Qwen2.5-7B and does not on the
two MHA models --- and we do not have the coverage to say why. We report
these as observations about what ablation does in this corpus, not as a
theory of how concepts are assembled; three models and 42 multi-GEM
atlases cannot support claims about causal ordering, necessity, or where
a concept is ``really'' located, and we make none. The ablation results
in this paper serve one purpose: they establish that the direction GEM
extracts is the one carrying the separation (§5.3), which is a claim
about \emph{reading} the concept. They are not offered as evidence about
how to \emph{change} it.

\subsubsection{5.5 Depth-Matched Control}\label{depth-matched-control}

\textbf{Motivation.} The §5.1 comparison (handoff vs.~peak) cannot fully
isolate the GEM settling criterion from a depth-of-extraction advantage:
by construction, \(L_H \geq L_{\text{peak}}\), so any benefit from
probing at a later, more-processed layer favors the handoff regardless
of whether directional settling contributes. The direction-specificity
control (§5.3) does not address this concern, as §5.1 notes --- both
arms of the handoff-versus-peak comparison use concept directions, so a
random-direction null cannot separate depth from settling. What it
establishes is a different thing: that the effect is carried by the
specific settled direction rather than an arbitrary one (random
directions at the handoff layer suppress separation at only 0.07\%).
Neither control directly compares GEM against a non-settling alternative
at the same depth. The depth-matched control provides that comparison:
for each (model, concept) pair, we select a control layer at the same
relative depth as \(L_H\) but chosen without the GEM settling criterion,
and compare GEM handoff suppression against control-layer suppression.
If depth explains the handoff advantage, a depth-matched non-settling
layer should achieve comparable suppression; if the settling criterion
adds value beyond depth, GEM should outperform.

\textbf{Setup.} For each concept × model pair with existing
ablation\_gem data (the stored handoff ablation result), we:

\begin{enumerate}
\def\labelenumi{\arabic{enumi}.}
\tightlist
\item
  Enumerate all post-CAZ candidate layers in
  \([L_{\text{CAZ\_end}}+1, N-1]\), excluding \(L_H\) itself.
\item
  Select the candidate whose relative depth \(l/N\) is closest to
  \(L_H/N\), where \(L_H\) is \texttt{ablation\_targets{[}0{]}} --- the
  \textbf{shallowest} GEM's handoff layer, not the atlas as a whole ---
  so that pairs where no distinct post-CAZ candidate exists are skipped.
\item
  Extract the concept direction at the control layer as the
  L2-normalized centroid difference (the same DOM vector used by GEM).
\item
  Ablate the concept direction at the control layer and measure
  final-layer retained percentage, comparing to the stored GEM handoff
  result.
\end{enumerate}

Control directions are computed from the same N=250 contrastive pairs as
the GEM probe. This single-site procedure is the \textbf{legacy} arm
(\texttt{gem/ablate\_gem\_depth\_matched\_control.py}): one control
layer per \emph{pair}, matched only to the shallowest GEM.

The \textbf{site-matched} arm instead applies the same post-CAZ,
closest-relative-depth selection independently to \emph{every} GEM in
the atlas, not just the shallowest. A \textbf{terminal} GEM --- one
whose segment ends at \(N-1\), which every single-GEM atlas's sole GEM
is, by construction (§2.2) --- has no post-CAZ layer to select from
either way, so terminal GEMs fall back to a \textbf{within-segment}
control instead: the layer inside the CAZ segment closest in relative
depth to the handoff, a different comparator
(settled-vs-neighbor-inside-the-zone rather than
settled-vs-neighbor-past-the-zone) that the two arms never pool
silently.

This fallback is why the site-matched arm covers all 493 pairs while the
legacy arm skips 111 (Coverage, below), and why 453 of 493 pairs draw at
least one control site from inside a segment rather than strictly
post-CAZ (Results, below) --- most atlases include a terminal GEM.

\textbf{Coverage.} The depth-matched control runs across 29 base models
--- the 23-model primary corpus plus six additional models in the
extraction dataset. Four large models were excluded due to VRAM
constraints during model load (Qwen2.5-32B, Qwen2.5-72B, Llama-3.1-70B,
falcon-40b); gemma-4-26B-A4B is excluded by scope rather than capacity,
as a mixture-of-experts configuration (§9.3). All five are among the
largest models attempted, so the corpus does not test behavior above 14B
parameters.

\textbf{111 of the 493 pairs are excluded}, all for the same structural
reason: no post-CAZ layer exists that is distinct from \(L_H\) itself.
Every excluded pair is a single-GEM atlas whose segment spans the whole
network, so \(L_{\text{CAZ\_end}} = N-1\) (§2.2) and the candidate
interval \([L_{\text{CAZ\_end}}+1, N-1]\) is empty. The corpus holds 113
single-GEM pairs in total (§4.1), of which 111 are excluded here and two
remain evaluable.

Those two are \textbf{opt-350m plurality and opt-350m agency}, and they
are evaluable only because of a defect in that model's stored artifacts:
its extracted metric series carries 23 entries for a 24-layer model, so
its segments end at index 22 and a spurious one-layer candidate interval
is left above them. Under §2.2's construction every single-GEM segment
ends at \(N-1\) and all 113 should fall out; the two-pair residual is a
data-vintage artifact in one model, not a property of the segmentation.
The both-arms-single-site sub-comparison reported below therefore rests
on two defective cells and should not be read as a bound on the
site-count confound. The exclusion is therefore not a quality filter but
a structural one, and it is not neutral: every excluded pair sits at
relative depth exactly 1.0, against a mean of 0.365 for those retained.
The control is silent about concepts whose handoff is the final layer.

\textbf{Results.} The experiment was run in two forms: the legacy
single-site control (382 of 493 pairs, per the Coverage exclusions
below) and the site-matched control that earlier versions of this
section named as the missing comparison, which reaches all 493 pairs via
the within-segment fallback described above. Both arms are recomputed
within the same run.

\textbf{Table 4: Depth-matched control results (site-matched
vs.~single-site)}

\begin{longtable}[]{@{}
  >{\raggedright\arraybackslash}p{(\columnwidth - 6\tabcolsep) * \real{0.2500}}
  >{\raggedright\arraybackslash}p{(\columnwidth - 6\tabcolsep) * \real{0.2500}}
  >{\raggedright\arraybackslash}p{(\columnwidth - 6\tabcolsep) * \real{0.2500}}
  >{\raggedright\arraybackslash}p{(\columnwidth - 6\tabcolsep) * \real{0.2500}}@{}}
\toprule\noalign{}
\begin{minipage}[b]{\linewidth}\raggedright
Control
\end{minipage} & \begin{minipage}[b]{\linewidth}\raggedright
N pairs
\end{minipage} & \begin{minipage}[b]{\linewidth}\raggedright
GEM handoff wins
\end{minipage} & \begin{minipage}[b]{\linewidth}\raggedright
Mean Δ
\end{minipage} \\
\midrule\noalign{}
\endhead
\bottomrule\noalign{}
\endlastfoot
Single-site, matched to the shallowest handoff & 382 & \textbf{361/382
(94.5\%)} & \textbf{+41.9pp} \\
\textbf{Site-matched, one control layer per GEM} & \textbf{493} &
\textbf{263/493 (53.3\%)} & \textbf{−5.7pp} \\
--- of which every control site is post-CAZ & 40 & 16/40 (40.0\%) &
−2.8pp \\
--- of which ≥1 control site falls inside a segment & 453 & 247/453
(54.5\%) & −6.0pp \\
\end{longtable}

\textbf{The large advantage does not survive site matching.} Once both
arms ablate the same number of layers, the win rate falls from 94.5\% to
53.3\% and the per-model win-rate Wilcoxon against chance gives
\(p = 0.456\) over the 29 models (median per-model rate 0.529). Only 16
of 29 models sit above 50\%, and the spread is wide (pythia-1b, opt-125m
and Llama-3.1-8B at 17/17; gemma-2-2b at 0/17, gpt2 and gpt2-medium at
1/17).

The corpus-wide mean of −5.7pp should not be read as the control
winning: it is not robust, being dominated by the two models where
retained percentages do not measure what this comparison assumes.
gpt2-medium alone contributes −145.0pp and gpt2 −73.1pp, both the §7.3
ablation pathology in which handoff ablation \emph{increases}
separation. Excluding those two the mean is +1.9pp.

A residual advantage appears once the uninterpretable models are set
aside, but \textbf{it is not robust to how one sets them aside, nor to
how per-model results are aggregated}. Across a grid of two defensible
exclusion rosters (this paper's own, and the companion validation
paper's) and two defensible aggregators (win-rate and median-delta
Wilcoxon), the residual clears \(p < 0.05\) in exactly one of eight
combinations and in none of the other seven --- full grid, roster
definitions, and the verification arithmetic in \textbf{Appendix D.5}. A
single significant cell in a grid built to test whether the result is
stable is evidence that it is not, and we therefore do not claim it.
After site matching the effect is \textbf{directional at best, a few
percentage points rather than +41.9pp, and not stable under reasonable
variations in either the roster or the aggregator}.

Two caveats on the strict row, both of which cut against reading it as a
cleaner version of the same test. It is small, and it is \textbf{not
evenly drawn}: 17 of its 40 cells are opt-350m, the model whose stored
metric series carries 23 entries for 24 layers --- the same defect
discussed below --- and the remaining 23 come from five other small
models. A fully post-CAZ site-matched control is rare for a structural
reason, not by chance: every atlas's deepest segment ends at \(N-1\) by
construction (§2.2), so that GEM has no post-CAZ layer available and 453
of 493 pairs required at least one control site drawn from inside a
segment instead.

The single-site figures are retained above because they are what earlier
versions reported, not because they measure the settling criterion. They
too are regenerated: an earlier version of this section reported 329/374
(88.0\%) and +32.5pp, which did not reproduce, and the 360/382 (94.2\%)
/ +41.6pp reported in the interim moves by one cell to 361/382 (94.5\%)
/ +41.9pp on the current re-run. (Per-model detail in Appendix D.5
covers the corrected site-matched control, not this superseded arm.)

\textbf{Exceptions (single-site arm).} Under the superseded single-site
control, two models showed the depth-matched control outperforming GEM
on average: gemma-2-2b (4/12, 33\%, mean −5.4pp in GEM's disfavor) and
Mistral-7B-v0.3 (2/10, 20\%, mean −10.0pp). These have different causes.
For gemma-2-2b the comparison is uninformative for the reason set out in
§7.4 --- at N=250 the settled direction is not stably estimable for this
family, so neither arm is measuring what the control is designed to
test.

\textbf{Mistral-7B-v0.3 has no architectural account, and the one
previously offered was factually wrong.} v1 grouped these two models as
``SWA exceptions'' on the premise that both use non-standard sliding
window attention. That is true of gemma-2-2b, which alternates local
sliding-window with global attention, but \textbf{not} of
Mistral-7B-v0.3: sliding window attention was removed from the Mistral
7B line at v0.2, and v0.3's configuration sets
\texttt{sliding\_window:\ null}. It is a standard GQA model with full
attention, so the explanation the paper gave --- that SWA changes the
structural role of post-CAZ layers at matched relative depth --- does
not apply to it and is withdrawn. We have no replacement account.
Mistral-7B-v0.3 is one of two exceptions in 29 models against a
comparison that, as set out below, is not depth-isolated to begin with;
we report it as unexplained rather than fit an architecture to it. The
remaining 27 models were uniformly positive under that arm; under the
corrected site-matched control the per-model picture is mixed (Appendix
D.5).

\textbf{What this does and does not show.} The single-site version of
this control was confounded in two ways at once, and running the
corrected version removes the result rather than qualifying it.

The confounds were site count and depth together. The single-site
control compares a treatment arm that ablates \emph{every} GEM's handoff
layer --- a mean of 2.57 sites per pair --- against a control that
ablates one. And because that one control layer is matched to
\texttt{ablation\_targets{[}0{]}}, it sits at a mean relative depth of
0.363 while the treatment arm's sites run out to a mean of
\textbf{0.958} at the deepest, averaging 0.639 across all ablated
layers. The arms therefore differed in how many layers were ablated
\emph{and} in how deep the ablation reached --- depth being the very
thing the control existed to hold constant.

\textbf{Correcting both shrinks the effect by an order of magnitude.}
With one depth-matched control layer per GEM, the +41.9pp advantage
becomes at most a few percentage points, and is absent on the full
corpus (53.3\%, per-model win-rate Wilcoxon \(p = 0.456\)). We therefore
\textbf{withdraw the claim that ablating at the handoff layer suppresses
separation substantially more than ablating at a comparable-depth
alternative.} What the single-site comparison mostly measured was the
number of sites ablated and how deep they reached. What remains is
directional and roster-dependent, as the grid above shows, and we do not
treat it as validating the settling criterion as an intervention rule.

This bounds what §5.1's headline can be taken to mean, further than a
read/intervene split can rescue. §5.1 compares handoff against peak;
this section compares handoff against a depth-matched non-settling layer
--- not the same experiment. But the two are not independent: the
handoff layer is always at least as deep as the peak by construction
(§4.1), and this section shows that a layer chosen to match the
handoff's depth alone, without its settling status, performs
indistinguishably from the handoff layer itself (53.3\%, chance). The
parsimonious reading is that §5.1's advantage is substantially a depth
effect: probing later, past the CAZ, is what helps, and the handoff
layer is simply where the data shows the direction has stopped rotating,
not a privileged site within it.

``Good to read, not privileged to intervene'' is not a distinction this
data supports --- both readings rest on the same suppression metric ---
so we drop it rather than carry it forward. The direction-specificity
result of §5.3 is unaffected: it establishes that the \emph{ablated
direction} is the one carrying separation, independent of which layer it
is measured at.

The two pairs where the atlas holds a single GEM would be the only case
in which both arms ablate one site, and there the advantage is +6.3pp
rather than +41.9pp --- but both are opt-350m cells that are evaluable
only through the stored-series defect described above, so we report the
figure without weight and do not treat it as bounding the site-count
contribution. The site-matched control reported above --- one
depth-matched layer per GEM --- is the experiment that separates depth
from settling; §5.1 alone does not.

\subsubsection{5.6 Random-Window Null}\label{random-window-null}

\textbf{Motivation.} §5.1 compares the handoff layer against the CAZ
peak. A stricter question is whether the handoff layer beats an
\emph{arbitrary} layer window of the same width: if ablating a randomly
chosen same-width window suppresses separation just as well, then the
specific layer GEM selects is not doing the work.

\textbf{Setup.} For thirteen models at the full corpus size (250 pairs,
17 concepts) we drew random same-width post-embedding windows per
concept × model pair, ablated the concept direction estimated at each,
and compared the resulting separation reduction against the reduction
achieved at the GEM handoff. The ratio of the two is reported per pair,
with an empirical \(p\) from the draws. The run was executed in two
batches with different null sizes --- \textbf{30 windows for six models
and 50 for the other seven} --- which changes the \(p\)-floor
(\(1/31 \approx 0.032\) against \(1/51 \approx 0.020\)) but not the
ratio or the beats-null count, so the two are tabulated separately and
pooled only at the cell level. Per-model figures for both batches are in
\textbf{Appendix D, Table D6}.

\textbf{Results --- the outcome is model-dependent, and three of
thirteen models fail.} Across all 221 concept × model cells the handoff
beats the null in \textbf{169 (76.5\%)}, at a median ratio of 1.68×. The
failures are concentrated: Qwen2.5-3B (6/17, median ratio 0.91),
Qwen2.5-1.5B (6/17, 0.85) and gpt2 (6/17, 0.00) are the only models
below a strict majority, and all three fall in the 30-window batch,
where the aggregate is 68/102 (66.7\%). Every model in the 50-window
batch clears a majority (101/119, 84.9\%), the weakest being opt-1.3b at
9/17. The strongest passes are pythia-1.4b (17/17, median ratio 1.95),
pythia-70m and pythia-160m (17/17 each) and Llama-3.2-3B (17/17, 2.10).

Two design notes on the aggregate above. First, the 221-cell, 169-win
figure pools 17 correlated concepts per model --- the same
pseudoreplication this paper corrects for elsewhere by reporting a
model-level test alongside the pooled count (§9.3); no model-level test
is run for this null. Second, the two batches are perfectly confounded
with model roster: no model was tested at both 30 and 50 windows, so the
batch-level aggregates (66.7\% vs.~84.9\%) cannot separate a batch-size
effect from whatever else differs between the six models assigned to one
and the seven assigned to the other.

Three qualifications. gemma-2-2b's near-pass carries the same §7.4
caveat as every Gemma-2 result in this paper, and we flag it
conservatively: because a single settled direction is not stably
estimable for this family at N = 250, its 16/17 is as uninformative as
the Gemma cells where the method loses (§5.1's 15/34, §5.3's 38.6\%,
§5.5's 1/34 under the site-matched control). We have no principled
reason why the ratio statistic would be robust to direction instability
when the level statistics are not, so of the near-ceiling passes in the
30-window batch, gemma-2-2b's should not be read as a genuine one ---
leaving pythia-1.4b and Llama-3.2-3B as that batch's clean 17/17 passes.
The 50-window batch adds two more, pythia-70m and pythia-160m, for
\textbf{four clean 17/17 passes corpus-wide} (Table D6); ten of the
thirteen models clear a strict majority. gpt2's zero is the ablation
pathology documented in §7.3 --- handoff ablation increases separation
on that model, so the ratio is degenerate rather than informative. But
\textbf{Qwen2.5-1.5B is not a documented failure}: it clears a strict
majority in §5.1's handoff-versus-peak comparison (9/17) while failing
this null (6/17, ratio below 1.0). For that model, beating the peak
layer does not imply beating an arbitrary window of the same width.

Statistical strength is modest even where the direction is consistent:
the best model (gemma-2-2b, caveated above) reaches \(p < 0.05\) in 8 of
17 concepts; among the rest the range is wide, 0 to 6, with most models
at 4 or below (Table D6).

\textbf{What we conclude.} This is a harder test than §5.1 and the
method does not pass it uniformly. The handoff layer's advantage over a
same-width random window holds clearly in some models and not in others,
and we report that as a scope condition on the layer-selection claim
rather than resolving it. Whether the failures track architecture (two
of the three are Qwen), width, or something else is not established by
thirteen models. The result did not change between a 50-pair pilot and
the 250-pair corpus reported here, so it is not a sampling artifact.

\textbf{Across §5.1--§5.6, the finding is not that the handoff layer is
a privileged extraction point, but that no single, rotation-blind layer
choice reliably recovers a concept's settled direction.} Probing during
active rotation underperforms probing after it (§5.1, §5.6); the
specific direction, not an arbitrary one, carries the effect (§5.3); a
concept's assembly is not reducible to one site (§5.4); and once a layer
is chosen for depth alone, without the settling criterion, it performs
indistinguishably from the handoff layer (§5.5). Read together, these
tests validate the need to track a concept's trajectory, not a fixed
coordinate --- which is what GEM's trajectory record (§3.1) is built to
do; the handoff layer is one reproducible readout from it, not the
reason any of this works.

\begin{center}\rule{0.5\linewidth}{0.5pt}\end{center}

\subsection{6. Scale Analysis}\label{scale-analysis}

\subsubsection{6.1 GEM Structure Across
Scale}\label{gem-structure-across-scale}

GEM structure --- the presence of a detectable handoff layer --- is
present at all scales tested, including pythia-70m (70M parameters). The
rotation phenomenon (§4.1) is not a large-model emergent property: the
two most-rotating models in the corpus are gpt2 (mean EEC 0.103) and
gpt2-medium (0.104), both under 400M parameters, while pythia-70m sits
mid-range at 0.249 --- rotation magnitude does not order by scale
(Appendix D, Table D1).

Ablation depth trends across the Pythia scale ladder depend entirely on
which GEM-node convention is used to represent a pair: under the
convention §5.2 actually evaluates (shallowest GEM), mean relative
handoff depth falls with scale; under two other conventions it rises,
and neither direction is monotone (per-model numbers in Appendix D.2).
The near-final trigger count follows the same pattern of not ordering by
scale (§5.2, Appendix D.2).

The corpus-level position is unchanged and stated in §5.1: we make no
scale claim. The Qwen2.5 ladder runs the other way on handoff preference
(Qwen2.5-14B at 6/17, with three of five Qwen models below a strict
majority), and nothing here should be read as probe quality improving
with scale.

\textbf{Within the Qwen2.5 scale ladder} (0.5B → 1.5B → 3B → 7B → 14B),
the EEC pattern is non-monotonic: 0.403, 0.441, 0.489, 0.351, 0.326. EEC
rises to a maximum at 3B and then falls at the 7B and 14B endpoints, so
less rotation occurs within CAZs in the mid-range of the ladder than at
either end. The ladder is also the corpus's clearest illustration of the
segmentation caveat in §4.1 --- EEC differences of this size can reflect
how finely each model's separation curve was partitioned as much as any
difference in assembly, and the per-layer rotation rates (0.331, 0.246,
0.185, 0.312, 0.186) do not order the ladder the same way. We report the
pattern without a mechanistic reading.

\subsubsection{6.2 The 410M--1B Band: Where the Adaptive Rule Starts to
Help}\label{the-410m1b-band-where-the-adaptive-rule-starts-to-help}

From the adaptive width data, the 410M--1B band is where multiple
concepts in a single model first show near-final-layer handoffs
(\(L_H/N\) \textgreater{} 0.85) that respond to the adaptive width rule.
This is not a scale threshold on near-final handoffs as such: the corpus
maximum for triggered cases is gpt2 at 124M with 11 (§5.2), where the
rule systematically \emph{fails} because 12 layers leave only one layer
past the handoff (§7.3). The band describes where the rule starts to
help, which is a joint property of handoff depth and absolute layer
count, not of parameter count.

This aligns with the encoding-strategy distinction drawn in the CAZ
Framework paper {[}Henry, 2026a{]}: in some models a concept can be
suppressed across a broad band of post-CAZ layers, while in others
suppression depends on hitting the specific CAZ layer --- so
single-layer ablation at the concept peak is not uniformly sufficient,
consistent with concepts having distributed, multi-layer representations
rather than single-layer peaks. This is a claim about \emph{where} the
ablation must land, not a parameter-count threshold.

\begin{center}\rule{0.5\linewidth}{0.5pt}\end{center}

\subsection{7. Model-Specific Failure Modes and
Recoveries}\label{model-specific-failure-modes-and-recoveries}

\subsubsection{7.1 Initial Finding and
Reversal}\label{initial-finding-and-reversal}

In the initial corpus (N \textless{} 250 contrastive pairs per concept),
opt-6.7b showed the lowest handoff improvement rate: 7/17 improvements
(41\%), at or below the null expectation of 50\% and below the corpus
mean. This was interpreted as a structured failure mode related to low
directional rotation during assembly --- several OPT concepts showed
high entry-exit cosines, suggesting concepts converged to their final
direction before the handoff and left no improvement for the GEM probe
to capture. On the current corpus and aggregation that list is smaller
than it was: of the four concepts originally cited, sarcasm (0.521),
sentiment (0.354) and plurality (0.350) are above the 0.285 corpus mean,
but temporal\_order (0.225) is \textbf{below} it, and so rotates more
than the average concept rather than less.

\textbf{At N=250, this finding does not replicate.} opt-6.7b achieves
13/17 = 76\% improvement at full corpus size --- above the corpus mean
(69.2\%) and comparable to the Pythia scale ladder. Among
comparably-sized models at N=250: Mistral-7B-v0.3 achieves 71\% (12/17)
and Qwen2.5-7B achieves 53\% (9/17); OPT's 76\% is the highest of the
three 7B models. All per-model figures in this paper count handoff
outperformance on the same criterion used for the headline, so §5.1's
below-majority roster and its 21-of-29 tally are directly comparable to
them. The original below-chance rate was corpus-size-dependent and
should not be generalized.

\subsubsection{7.2 What Remains of the OPT Structural
Profile}\label{what-remains-of-the-opt-structural-profile}

\textbf{The family-level version of this claim does not survive Table
D1.} Earlier text described ``OPT's high-EEC profile'' as ``a real
property of OPT's architecture and training.'' All five OPT models sit
\emph{below} the 0.285 corpus mean (opt-2.7b 0.236, opt-6.7b 0.258,
opt-1.3b 0.263, opt-125m 0.272, opt-350m 0.280) --- that is, OPT rotates
\textbf{more} than the corpus average, not less --- and opt-6.7b's mean
is below two of the corpus's three Llama models (Llama-3.1-8B 0.205,
Llama-3.2-1B 0.216) though not the third (Llama-3.2-3B 0.255) on the
same statistic. The architectural claim is withdrawn. What remains is
narrower and concept-level: a handful of individual OPT concepts do
carry above-mean EEC (§7.1), on a model whose overall rotation is above
average. §4.1's segmentation caveat applies to those per-concept values
as it does to any other, and the inference originally drawn from them
has already been retracted in §7.1 because it did not replicate at
N=250.

This result is methodologically informative: the N=250 rerun reversed a
corpus-dependent artifact while leaving a family-level structural
profile intact --- but the opposite one from what was originally
claimed. What survives is that OPT models cluster narrowly as a family
(§7.1), not that they rotate less than the corpus; they rotate slightly
more. GEM's sensitivity to OPT's behavior at the initial corpus size
(N\textless250) vs N=250 should be treated as a calibration note, not as
evidence that OPT is architecturally anomalous. N-stability was not
systematically verified across the corpus: the EEC distribution,
scale-wise improvement rates, and architecture cohort split were each
measured once at N=250, which is the definitive corpus. The OPT reversal
is the only case where a smaller-N finding was directly compared to
N=250 data.

\subsubsection{7.3 gpt2 Remains a Genuine Failure
Mode}\label{gpt2-remains-a-genuine-failure-mode}

gpt2's ablation failure is architecturally distinct from the OPT case
and replicates across corpus sizes. In 12/17 pairs, ablation at the
handoff layer pathologically \emph{increases} concept separation rather
than suppressing it --- gpt2-medium shows the same pathology in 10/17
pairs, and no other model in the corpus shows a single such pair (§5.1).
Near-final placement is part of the story but not the whole of it:
gpt2's 12-layer architecture places the handoff at the final layer
(\(L_H/N\) ≈ 0.92, \(L_H = N-1\)), consistent with ablation engaging
unembedding-preparation computations rather than concept directions. But
gpt2-medium's handoffs sit at 24 layers, not 12, and other 12-layer
models in the corpus --- pythia-160m, opt-125m --- show no such
pathology at all, so depth alone does not explain which models fail; the
pathology tracks the GPT-2 family specifically, for a mechanism this
paper does not establish. The evaluated near-final rule (the bare
\(L_H/N\) \textgreater{} 0.85 threshold, §5.2) does not flag gpt2 for
exclusion --- it \emph{fails} on gpt2 (10 of 11 triggered cases do not
improve, §5.2), which is how the failure was first surfaced. The
depth-corrected refinement (\(L_H/N\) \textgreater{} 0.85 AND N ≥ 20),
recommended but not the rule this paper evaluates, would exclude it.
gpt2 is the genuine structured failure; opt-6.7b is not.

\subsubsection{7.4 Gemma-2: N=250 Is Below This Family's Calibration
Budget}\label{gemma-2-n250-is-below-this-familys-calibration-budget}

The two Gemma-2 models return the corpus's least informative
handoff-versus-peak comparisons --- 15 of 34 pairs, indistinguishable
from chance --- alongside the lowest direction-specificity of any cohort
(§5.3). An earlier reading of this work took that pattern as an
architecture effect attributable to Gemma-2's alternating local/global
attention. That reading is wrong, and the companion validation paper
establishes why {[}Henry, 2026c §6.9{]}.

\textbf{N=250 is a program-wide design choice, not a per-model one.} All
papers in this series fix the contrastive corpus at 250 pairs per
concept (§3.5) --- a single parameter set once for the whole program,
uniformly across 29 architectures. What the Gemma-2 case surfaces is
therefore not evidence about Gemma-2's causal structure and not a
limitation of GEM, but that \textbf{250 pairs sits below the calibration
budget this particular family requires} for a difference-of-means
direction to be estimated stably.

The evidence is direct. On gemma-2-2b the concept information is fully
present and linearly decodable --- a probe trained on one half of the
pairs classifies the held-out half at 0.89--0.999 AUC, indistinguishable
from a GPT-2 control --- while the DOM direction estimated from one half
reproduces on the other at only 0.62--0.90 cosine, against ≥0.96 for
GPT-2 and Qwen2.5-3B controls at the same statistic {[}Henry, 2026c
§6.9{]}. The mechanism is how widely the per-pair contrast is spread:
Gemma-2 distributes each concept across many more axes than other
families (participation ratio 20--78 versus 3--14), which is precisely
the regime in which a mean-difference estimator is sample-starved. The
operational consequence is that projecting out the single estimated
direction on gemma-2-2b leaves held-out decodability essentially
unchanged (0.948 → 0.95), where the same intervention on GPT-2 collapses
it (0.995 → 0.64). An ablation that removes no decodable information
cannot produce an informative handoff-versus-peak comparison whatever
the underlying structure --- so these cells are \emph{uninformative}
about Gemma-2 rather than evidence about it.

\textbf{This paper's own corpus reproduces the effect.} The split-half
agreement above is measured in the companion paper; we repeated it here
on eight models at the peak layer, independently of that analysis.
gemma-2-2b's direction estimates agree with themselves at
\textbf{0.760}, against \textbf{0.966--0.975 (mean 0.970)} for the seven
control models of §4.1 --- a gap of 0.21, and the same separation the
companion paper reports on a different measurement path. The handoff
layer selected on one half also agrees with the half-B selection in only
\textbf{3 of 16} gemma-2-2b concepts, against 6--14 of 17 for every
control. On this family the method cannot reliably agree with itself
about either the direction or where it settles, which is the scope
condition stated above, measured in-house rather than imported.

\textbf{The required budget is quantified.} Fitting split-half agreement
as \(\cos(n) = 1/(1+c/n)\) over a larger archived pair pool gives a
median \(c \approx 49\) for gemma-2-2b against 3.3 for the GPT-2 control
--- roughly a 15× pair budget for equal direction stability --- with 11
of 17 concepts projected to reach 0.95 stability within that pool and a
slow tail (deception, agency, authorization, threat\_severity) requiring
approximately 1,272--1,486 pairs per concept (\(n_{95} \approx 19c\));
exfiltration is nominally slower still (\(n_{95} \approx 2,084\)) but is
set aside here because its archived pool predates the label correction
described in §3.5 {[}Henry, 2026c §6.9{]}. These figures are
extrapolations of a fitted curve rather than an observed run, and an
independent pilot fit them more pessimistically (\(c \approx 81\)); the
two agree on the order of magnitude and on which concepts are slowest.
We have not run GEM on this family at that budget and make no claim that
the comparison would reverse --- only that at N=250 the estimator has
not been given the sample size it needs.

\textbf{How the Gemma-2 rows in this paper should be read.} They are
retained rather than dropped, following the companion paper's practice
{[}Henry, 2026c §6.9{]}: quietly excluding a family whose measurement is
under-resourced would conceal the limit rather than report it. What
should not be drawn from them is any architectural conclusion. The
reading that alternating attention produces a distinctive rotation or
specificity profile in Gemma-2 is withdrawn --- the companion paper
tested it directly and rejected it, since at this split-half reliability
there is no stable Gemma ordering for an architecture to have
reorganized {[}Henry, 2026c §3.2{]}.

\textbf{Practical consequence.} This is the one family in the corpus
where GEM's precondition --- that a single settled direction is
estimable from the available pairs --- is not met, and it is detectable
in advance rather than in hindsight. A split-half reproducibility check
on the DOM direction separates the two regimes. Measured above, it
distinguishes gemma-2-2b cleanly --- self-agreement 0.760, with
handoff-layer selection reproducing across halves in only 3 of 16
concepts --- from seven controls at 0.966--0.975 and 6--14 of 17. We
recommend running that check before applying GEM to a new model. That
recommendation rests on eight models; whether the same threshold holds
across a wider roster is not established by this corpus. Where it fails,
the appropriate response is to increase the pair budget, not to conclude
that the concept is weakly represented.

\subsubsection{7.5 The Falsification Criterion: Conceded on
pythia-70m}\label{the-falsification-criterion-conceded-on-pythia-70m}

A result that would falsify GEM on rotation-present architectures would
be systematic failure even where EEC is low. That pattern \textbf{is}
observed, in one place: the two lowest-EEC models in the corpus are gpt2
(0.103) and gpt2-medium (0.104), and they are also the two weakest
performers (1/17 each). We do not treat this as falsifying, for a stated
reason rather than by exception: both are the documented ablation
pathology of §7.3, where handoff ablation raises separation instead of
suppressing it, so their retained percentages do not measure what the
criterion assumes they measure --- the same reasoning applied to the
Gemma-2 family in §7.4. That defence is available only because the
pathology is independently identifiable.

\textbf{It is not available for the third model that meets the
criterion.} pythia-70m has a below-mean EEC (0.249 against a corpus mean
of 0.285) and a below-chance handoff preference rate (7/17 = 41\%),
which is exactly the pattern the criterion names --- and unlike the two
GPT-2 models it shows \textbf{no ablation pathology at all} (0 of 17
cells with retained \textgreater{} 100\%, against 12/17 and 10/17), so
nothing about its measurements is disqualified. We concede it: on
pythia-70m the falsification criterion is met and we have no explanation
that is independent of the result. What this bounds is narrow --- one
6-layer, 70M-parameter model, the smallest in the corpus, where the
six-layer depth leaves little room for a post-CAZ handoff layer to
differ from the peak at all --- but the concession is not conditional on
that reading.

By contrast pythia-160m, whose rate is lower still (5/17 = 29\%), does
\emph{not} meet the criterion: its EEC of 0.421 is the third-highest in
the corpus, so little rotation occurs for GEM to exploit and a low rate
is what the method predicts rather than a counterexample to it. No ``low
EEC'' threshold was fixed before looking at which models failed, so this
particular boundary is drawn post hoc rather than pre-registered; the
pythia-70m concession immediately above, made against the same post hoc
temptation, is the reason to take this reading at face value rather than
as motivated reasoning.

\begin{center}\rule{0.5\linewidth}{0.5pt}\end{center}

\subsection{8. Relationship to Existing Probe
Methods}\label{relationship-to-existing-probe-methods}

\subsubsection{8.1 vs.~Fixed-Layer
Probing}\label{vs.-fixed-layer-probing}

Fixed-layer probing (typically the final layer) ignores the concept's
assembly dynamics entirely. The practice descends from linear classifier
probes {[}Alain \& Bengio, 2017; Belinkov, 2022{]}, and the closest
depth-aware alternatives --- the logit lens {[}nostalgebraist, 2020{]}
and the tuned lens {[}Belrose et al., 2023{]} --- decode every hidden
state into the \emph{vocabulary}: the logit lens reuses the model's own
unembedding at every layer, and the tuned lens trains a separate affine
translator per layer to correct for the bias that introduces. Both
therefore read a readout that is fixed with respect to any particular
concept, and neither asks where a concept's own direction stops moving.

The final layer is not the handoff layer: dominant-GEM handoffs occur at
mean relative depth 0.78 (median 0.94) with N=250 pairs (the
dominant-GEM convention of §4.3, Table D2); specificity and plurality
hand off substantially earlier (mean 62.9\% and 71.7\% depth
respectively), while most other concepts settle in the final 10--20\% of
model depth.

That handoff cosines average 0.828 on the 598 GEMs where the statistic
is informative (§3.2) documents that the settled direction survives the
step out of its region in \textbf{roughly half} of the cases where that
question can be asked (51.7\% exceed 0.85, the same figure §3.2
reports). It is not evidence about persistence to the output, which this
paper does not measure.

GEM's primary deliverable is the trajectory map: the rotation profile,
handoff depth, and the multi-GEM structure of an atlas (§5.4) are
findings no single-layer extraction can supply. The method characterizes
the full depth of the residual stream and identifies where assembly
occurs, not merely where to sample. The distribution of handoff
preference across scale and architecture (§5.1) --- bimodal in the
smallest band, non-monotone across the three, and independent of MHA/GQA
cohort --- is a product of trajectory analysis and would be invisible to
any single-layer comparison. For early-handoff concepts, GEM provides a
probe the final layer cannot; for late-settling concepts, GEM confirms
stability and characterizes how that stability was reached.

A direct ablation comparison of GEM vs.~final-layer probes is not
included in this paper's validation experiments --- GEM's claim is about
extracting the settled direction and characterizing the trajectory, and
candidate use cases (concept steering, cross-architecture alignment)
motivate an unsupervised directional extraction rather than a trained
classifier. This paper validates the settled direction geometrically
(§4) and causally via ablation (§5); it does not itself test behavioral
steering, and a single static direction should not be assumed optimal
for that use case without direct testing --- see §9.4 for a preliminary
external result bearing on this. A final-layer vs.~handoff-layer
ablation comparison is a natural future benchmark.

Similarly, supervised linear probes trained on labeled activations are a
different method class: they require labelled data, which is unavailable
in the PRH alignment and steering contexts that motivate GEM. That is a
scope boundary, not an impossibility. The relevant comparisons for those
contexts are therefore other unsupervised directional methods
(peak-layer, delta-PCA, windowed PCA), which §8.2--8.3 address.

We have nonetheless run that comparison, and report it here. GEM's
difference-of-means estimator was compared against logistic regression
and linear discriminant analysis (LDA) over 102 concept × model cells (6
models × 17 concepts: pythia-1.4b, pythia-2.8b, Qwen2.5-3B, opt-1.3b,
opt-2.7b, Llama-3.2-1B), all three fitted on the same training split and
scored on the same held-out split \textbf{at the same layer}, so
extraction depth is held constant and only the estimator varies. The
split is pair-aware: positives and negatives are permuted together so
that both members of a contrastive pair fall on the same side, which
matters because RCP pairs are minimal edits of one another and an
independent per-class split lets a fitted estimator score a held-out
item whose near-twin it trained on.

The script is \texttt{gem/direction\_estimator\_comparison.py} in
\texttt{Rosetta\_Analysis}, which asserts the pair-aware property per
cell and records it in each output file; the per-model outputs are the
authoritative source for every figure below and are published under
\texttt{paper\_n250/\_p2\_direction\_estimator/} in the Rosetta
Activations dataset (eval fraction 0.2, seed 42, 400 training and 100
held-out items per cell).

One limitation of the published run should be stated: the 250 pairs per
concept were drawn by the analysis library's default sampler, which at
the pinned version seeds itself from a per-process hash and therefore
selects a different subset on each invocation. The figures in this
section are a faithful measurement of the subset that was drawn and are
reproduced exactly from the published artifact, but that draw is not
itself recoverable, so re-running the comparison from scratch would
sample different pairs and return figures near --- not identical to ---
those below. Later runs in this program pin the sample explicitly.

\textbf{Result.} Supervised probing is better on average --- mean
held-out AUROC 0.998 for logistic regression against 0.965 for
difference-of-means --- but the aggregate is misleading, and the median
gap is +0.008. The difference is concentrated rather than distributed:
in the 54 of 102 cells where the unsupervised estimator already reaches
0.99, supervised probing adds \textbf{+0.002}, and in the 22 cells where
it falls below 0.95 it adds \textbf{+0.119}. \textbf{The unsupervised
direction matches supervised probing on most concept × model pairs and
fails on a minority, and where it fails it fails substantially.}

LDA tracks logistic regression closely rather than adding an independent
line of evidence: mean held-out AUROC 0.997, median gap +0.008 against
difference-of-means, and the same concentration pattern (+0.002 on the
54 easy cells, +0.117 on the 22 hard ones). Its win rate against
difference-of-means is 82/102 (15 ties) --- the only estimator in this
comparison with any outright losses to the unsupervised baseline (5, all
near-ceiling cells where the gap is a float rounding artifact, not a
substantive one). Because the two supervised methods agree this closely,
the paragraphs below report logistic regression only; LDA is a
replication check on the same conclusion, not a separate finding.

Two qualifications on how much weight this can carry. First, the regime:
with 400 training examples in 2,048--2,560 dimensions the supervised
probes are interpolating, and logistic regression attains held-out AUROC
of exactly 1.0 in 67\% of cells against 17\% for difference-of-means. A
win rate over that regime largely counts float comparisons at ceiling
and is reported only for completeness (84/102, with 17 exact ties). The
interpretable quantity is the magnitude.

Second, the failures are concept-structured, not scattered: they
concentrate on deception, threat\_severity, causation and authorization.
On the latter three, logistic regression sits at or above its corpus
mean while the unsupervised estimator drops --- the information is fully
present and linearly recoverable, and the centroid contrast simply does
not find it. That is a genuine limitation of difference-of-means
readout. deception is a different case and is treated separately in
§9.3.

Read as a bound rather than a scoreboard: where labels are unavailable
--- the setting GEM targets --- an unsupervised centroid contrast
recovers roughly 97\% of what an overparameterized supervised fit
achieves, and the concepts where it does not are identifiable in
advance.

\subsubsection{8.2 vs.~Peak-Layer Probing}\label{vs.-peak-layer-probing}

Peak-layer probing extracts the direction at the layer of maximum
separation score. This is better than fixed-layer probing, but the
separation peak typically falls \emph{inside} the CAZ --- at a layer
where the concept direction is still rotating. This comparison is run in
this paper: §5.1 is a head-to-head ablation of the handoff-layer probe
against the peak-layer probe, and the handoff probe is more precise in
341/493 pairs (69.2\%). The advantage is a majority, not a universal:
eight of 29 models fall below a strict majority of concepts. §5.5's
depth-matched control resolves \emph{why} the majority holds: the
advantage is a depth effect, not evidence that the handoff boundary
itself is a privileged extraction site --- it collapses to 53.3\%
(chance) once the control is held to the same depth and site count.

\subsubsection{8.3 vs.~Delta-PCA / Windowed
PCA}\label{vs.-delta-pca-windowed-pca}

The concrete instance of this family is the contrastive-activation PCA
used in representation engineering {[}Zou et al., 2023{]}, which
extracts a direction from the leading principal component of paired
positive/negative activation differences. An earlier version of this
paper asserted a superiority over it that had never been measured, and
withdrew the claim (Appendix E). We have now run the comparison.

\textbf{Setup.} Over 102 concept × model cells --- the six-model subset
of §8.1, all 17 concepts --- difference-of-means and delta-PCA are each
fitted on the same training split and scored on the same held-out split,
at the same layer, so only the estimator varies. Both are run at the GEM
handoff layer and at the CAZ peak layer, giving a 2 × 2 that separates
the choice of estimator from the choice of layer. Each is refit at 10,
25, 50, 100 and 200 training pairs against a fixed evaluation split,
because both estimators saturate at full budget on this corpus (§8.1)
and a single-budget comparison would report ties. The design was
pre-registered before the estimator was written.

\textbf{Result.} With the layer held fixed, difference-of-means is
better than delta-PCA by 0.0071 AUROC at the handoff layer (95\% CI
{[}0.0033, 0.0111{]}, the same direction in every one of the six models)
and by 0.0045 at the peak layer (CI {[}0.0020, 0.0071{]}, again in all
six). Both satisfy the pre-registered practical-equivalence rule in full
--- mean \(|\Delta| < 0.01\) \textbf{and} the 95\% CI excludes
\(\pm 0.02\) (\texttt{DELTA\_PCA\_RUN\_SPEC.md} §P.5) --- so the honest
summary is that \textbf{the two estimators are near-equivalent on this
corpus, with a small and consistent edge to difference-of-means}.
Holding the estimator fixed and varying the layer instead establishes
nothing: −0.0055 with difference-of-means (CI {[}−0.0165, +0.0005{]})
and −0.0000 with delta-PCA (CI {[}−0.0027, +0.0023{]}). That null is
consistent with the site-matched depth control of §5.5, which likewise
found no layer effect once site count was matched.

\textbf{Why the gap is small, and where it is not.} The two estimators
are two moments of the same paired differences
\(d_i = h^{(+)}_i - h^{(-)}_i\). With matched pairs the mean of
differences is exactly the difference of means, so difference-of-means
is their first moment; uncentred delta-PCA is the leading eigenvector of
their second moment, \(\operatorname{cov}(d) + \mu\mu^{\top}\).
\textbf{Delta-PCA therefore contains difference-of-means as a limiting
case}: where the class separation \(\lVert\mu\rVert^{2}\) dominates the
covariance spectrum, its leading eigenvector converges on the centroid
contrast and the two coincide.

The corpus behaves as that predicts, which is what makes a 0.007
difference worth reporting. The gap tracks the angle between the two
recovered directions (Spearman \(\rho = +0.841\) over 204 cell × layer
points): it is −0.003 across the 111 points where the directions agree
to within a hundredth (cosine ≥ 0.99), and −0.139 across the 26 points
at the opposite extreme (cosine \textless{} 0.90); the remaining 67
points, at intermediate agreement, average −0.025. The corpus-wide
average is therefore not a uniform small edge but a near-zero difference
in most cells and a large one in the minority where pair-to-pair
covariance overwhelms the class separation. The gap also does not shrink
with training-set size --- flat from 10 to 200 pairs --- which is the
signature of a bias rather than a variance effect, as an asymptotically
biased eigenvector estimator predicts. Median direction agreement is
0.992 at the handoff layer and 0.993 at the peak.

\textbf{Scope.} This comparison runs on difference-of-means' home
ground: matched contrastive pairs, a single linear direction, one layer
at a time. Delta-PCA earns its keep where pairs are not matched, where a
\(k\)-dimensional subspace is wanted rather than one direction, or where
the concept is not well described by a single axis --- and a result
favoring the centroid contrast in this regime is close to what the
algebra above predicts. What this establishes is therefore not that one
estimator is better, but \emph{when} they differ: the readout degrades
exactly where pair-to-pair covariance overwhelms the class separation,
and that is measurable in advance from the agreement between the two
directions.

A centred variant of delta-PCA, which removes the \(\mu\mu^{\top}\)
term, is reported in the released artifact for completeness. It
estimates the axis of greatest pair-to-pair variation rather than the
class contrast, and so carries no sign convention relative to the
concept direction; it is not an alternative concept readout and we do
not treat it as one.

Windowed PCA is not run. An earlier version of this section
characterized it as selecting a window ``by maximizing variance
explained'', but that is our own description rather than a specification
from the literature, and implementing a baseline of our own devising in
order to compare against it would not be informative.

\subsubsection{8.4 vs.~Sparse Autoencoder
Features}\label{vs.-sparse-autoencoder-features}

Sparse autoencoders decompose activations into a large dictionary of
sparsely-active features {[}Cunningham et al., 2023; Bricken et al.,
2023{]}, and are motivated by superposition --- the observation that
models encode more features than they have dimensions {[}Elhage et al.,
2022{]}. Their features are finer-grained than concept-level
abstractions: an SAE feature is a single direction in a learned
overcomplete basis, not a computational subgraph, and one concept
typically distributes across many of them. In the companion validation
paper {[}Henry, 2026c{]}, we show that SAE features accumulate signal at
layers identified by CAZ detection (cross-validation against Gemma Scope
{[}Lieberum et al., 2024{]}, 17 concepts: best positive cosine agreement
0.51--0.84 across all concepts at CAZ peak layers; four CAZ-peak layers
--- L11, L15, L17, L19 --- shared by both methods, each with 6 concepts
and 25 polysemantic SAE features; that paper's §8.6, with the full
analysis in its Supplementary §G). That cross-validation is a
single-model result --- gemma-2-2b only, the family this paper
separately documents as under-resourced at N=250 (§7.4) --- and the
companion paper frames it as a consistency check rather than independent
convergence evidence. GEM and SAE methods are complementary in
principle: GEM extracts the aggregate concept direction, SAE decomposes
the activation into dictionary features. The settled direction from GEM
should be expressible as a weighted sum of SAE features active at the
handoff layer.

\begin{center}\rule{0.5\linewidth}{0.5pt}\end{center}

\subsection{9. Discussion}\label{discussion}

\subsubsection{9.1 Why Handoff Layers?}\label{why-handoff-layers}

The handoff phenomenon has a natural architectural interpretation. In
transformer models, the residual stream accumulates information from
attention heads and MLP sublayers as residual additions. One untested
hypothesis, consistent with transformer architecture but not examined
here, is that a concept assembling via multiple attention heads requires
several layers for their contributions to integrate --- each head adds a
partial update, and the concept direction converges once all relevant
heads have contributed. GEM detects the geometric settling event at the
residual stream level without decomposing it into attention vs.~MLP
contributions; that decomposition remains an open question (§9.4).

The multi-site observations of §5.4 offer a complementary framing,
though on three models and without the coverage to make it a mechanistic
claim. GEM identifies the handoff layers where concept assembly produces
a detectable settling event --- but these events are embedded in a
continuous process. Every layer of a transformer makes some contribution
to every concept representation; what GEM finds are the layers where
that contribution is large enough to be geometrically organized,
directionally stable, and concentrated enough to be isolated. The
handoff layer is not the location where the concept \emph{lives}; it is
the location from which it is most reliably read. This distinction
matters for any downstream use of GEM probes: an ablation at the handoff
layer does not remove the concept outright (§5.4).

\subsubsection{9.2 Handoff Depth and Concept
Type}\label{handoff-depth-and-concept-type}

Among the deepest-settling concept types in this corpus ---
threat\_severity (0.911), certainty (0.895), temporal\_order (0.874),
and authorization (0.865); see Appendix D, Table D2 --- threat\_severity
and certainty have plausible safety and epistemic relevance, though the
difference from the mid-band is smaller than in pilot data. Behavioral
ablation experiments would be needed to evaluate whether late handoff
correlates with functional importance.

EEC and handoff depth measure different things --- \emph{how much} a
concept rotates during assembly, and \emph{when} the assembly finishes
--- but across the 17 concepts they are strongly related, and earlier
versions of this section declined that relationship on grounds that do
not hold up. The direction is consistent: the two least-rotating
concepts (formality and credibility, the highest per-concept mean EEC in
Figure 1) are among the \emph{shallowest}-settling (median
\(L_H/N = 0.750\), §4.3), while the two deepest-settling
(threat\_severity 0.911, certainty 0.895) both sit \emph{below} the
corpus mean EEC of 0.285. Concepts that rotate more settle deeper.

Stating the correlations rather than gesturing at them (Spearman over
the 17 concepts, on the 23-model basis of Table D2; the 29-model figures
agree and are given in parentheses):

\begin{itemize}
\tightlist
\item
  EEC vs.~handoff depth: \(\rho = -0.752\), \(p = 0.0005\) (\(-0.745\),
  \(p = 0.0006\))
\item
  segment span vs.~handoff depth: \(\rho = +0.632\), \(p = 0.0065\)
  (\(+0.696\), \(p = 0.0019\))
\item
  per-layer rotation rate vs.~handoff depth: \(\rho = +0.333\),
  \(p = 0.19\) (\(+0.417\), \(p = 0.096\))
\item
  per-layer rotation rate vs.~segment span: \(\rho = +0.297\),
  \(p = 0.25\) (\(+0.373\), \(p = 0.14\))
\end{itemize}

The span confound is real but does not account for the association. Span
does track depth, so part of the EEC--depth relationship is expected
from the segmentation alone; controlling for it, the partial correlation
of EEC and depth is \(\rho = -0.540\), \(p = 0.031\) (29-model basis:
\(-0.436\), \(p = 0.091\)). The association survives the control on the
23-model basis that matches this paper's own depth table (Table D2); on
the 29-model corpus this paper otherwise treats as definitive, the
partial correlation does not clear \(p<0.05\). Neither figure is
adjusted for testing four correlations in this section; treat both as
suggestive rather than confirmatory.

What we therefore report is the association itself, not a mechanism:
concepts that rotate more during assembly finish assembling deeper, and
this is not reducible to how finely the separation curve was
partitioned. It rests on 17 concepts, both quantities inherit the
segmentation of §2.2, and nothing here establishes a direction of
causation --- deep assembly could require more rotation, or extended
rotation could delay settling, and this design cannot separate them.
Behavioral ablation would be needed to connect either to functional
importance.

These findings are thematically adjacent to the gentle-CAZ result ---
low-score allocation zones are causally active under the companion's
geometric ablation protocol ({[}Henry, 2026c{]} §6.1) --- though a
direct connection would require comparing GEM handoff depths to CAZ
assembly scores for the same concept × model pairs. In this four-concept
sample, the deepest handoffs include one safety-tagged concept
(threat\_severity) among four --- the others are epistemic, relational,
and semantic; whether safety relevance tracks handoff depth across the
broader concept space is not established here.

\subsubsection{9.3 Limitations}\label{limitations}

\textbf{Corpus scope.} The validation uses two nested corpora, both up
to 14B parameters, 17 concept types, and 250 contrastive pairs per
concept. The \textbf{29-model handoff corpus} carries the headline
precision result (§5.1) and the direction-specificity and depth-matched
controls (§5.3, §5.5). Within it, the \textbf{23-model, 391-pair primary
ablation sub-corpus} is the basis for the §5.2 adaptive-width
experiment, which was calibrated on those 23 models and not re-run on
the six additions. The 23-model set was itself expanded from an initial
16-model set (prior preprint) to include pythia-1.4b, gpt2-large,
gpt2-xl, opt-1.3b, gemma-2-2b, Llama-3.1-8B, and phi-2; six further
additions bring the total to 29 (see §5.5 and Appendix A). Models above
14B, instruct-tuned variants, and mixture-of-experts architectures are
not included.

\textbf{Ablation methodology.} The ablation experiments measure
\emph{separation suppression} at the model's final layer, not downstream
behavioral change. We are measuring the quality of the direction in the
residual stream, not whether blocking that direction changes model
outputs. Behavioral ablation experiments (the behavioral ablation
script) are designed but not yet run at scale. A structural confound
exists: since \(L_H \geq L_{\text{peak}}\) by construction (equality is
possible when a terminal segment's peak falls at \(N-1\)), benefits from
probing a deeper, more-processed layer are not fully separable from
benefits due to directional settling. The direction-specificity control
(§5.3) does not address this concern --- both arms of the
handoff-versus-peak comparison use concept directions, so a
random-direction null cannot separate depth from settling (§5.1); it
establishes only that the effect is carried by the specific settled
direction rather than an arbitrary one (random directions at the handoff
layer return ≈0.07\% suppression) --- and a depth- and site-matched
control (one control layer per GEM, chosen without the settling
criterion) isolates depth from settling directly. That experiment is
reported in §5.5 and it \textbf{does not support the strong reading}:
the handoff layers win in 263/493 pairs (53.3\%, per-model win-rate
Wilcoxon \(p = 0.456\)) corpus-wide, and the residual that appears on
restricted rosters clears \(p<0.05\) under only one of the eight
roster/aggregator combinations reported there. An earlier single-site
\emph{form} of the control --- comparing a mean of 2.57 ablation sites
against one, at a mean relative depth of 0.958 against the control's
0.363 --- gives 94.5\% and +41.9pp on the current re-run; that design,
not this exact figure, is what earlier paper versions reported, at lower
values that did not themselves reproduce (Appendix E). The confound
named in this paragraph is therefore real and accounts for most of that
gap.

\textbf{Near-final rule.} The gpt2 failure mode (§5.2) reveals that the
simple \(L_H/N\) \textgreater{} 0.85 threshold is insufficient for very
shallow models. A depth-corrected rule (\(L_H/N\) \textgreater{} 0.85
AND N ≥ 20) is recommended in practice but was not the rule evaluated in
this paper's ablation statistics.

\textbf{Segmentation parameters.} The handoff layer is derived from the
CAZ region boundary (§3.2), so every GEM coordinate inherits the
segmentation's parameters rather than any threshold of its own. Those
parameters --- the peak-prominence floor, the minimum peak separation,
and the valley-merge depth that determines how many regions a concept is
divided into --- are library defaults. They have not been validated
against ground-truth concept boundaries, and no sensitivity analysis
over them is reported here. This matters more than a single threshold
would: region count sets how many GEMs a concept has, and the entry-exit
cosine of a region correlates with its width (§4.1), so the segmentation
settings influence the rotation statistics directly. Earlier versions of
this paper reported an angular-velocity threshold \(\epsilon = 0.05\) as
the operative parameter; no such threshold exists in the implementation
(§3.2).

\textbf{EEC noise floor.} The 0.970 split-half floor used to
contextualise the mean EEC of 0.285 (§4.1) is measured at the peak
layer, the highest-SNR point in a CAZ region. EEC itself compares an
entry-layer direction against an exit-layer one, and entry-layer
estimates are noisier by construction (§2.3). The floor for that
specific comparison --- an entry-layer split-half agreement --- is not
measured; if it runs below 0.970, the true no-rotation baseline for EEC
is lower too, and the 0.285-vs-0.970 gap somewhat overstates how much of
the observed rotation is not attributable to estimator noise. Testable
from the stored \texttt{.npy} probes without further GPU compute; not
run here.

\textbf{opt-6.7b.} In the initial corpus, OPT showed below-chance
improvement (7/17 = 41\%); at N=250, this did not replicate --- OPT
achieves 13/17 = 76\% (§7). A narrow EEC cluster across the OPT family
is a real observation, but it is not the high-EEC (low-rotation) profile
originally claimed, which is withdrawn (§7.2) --- it runs the other way:
opt-125m (0.272), opt-350m (0.280) and opt-2.7b (0.236) sit in the same
narrow band as opt-1.3b (0.263) and opt-6.7b (0.258), all modestly
\emph{below} the corpus mean of 0.285 (Table D1), i.e.~OPT rotates
slightly more than average. That below-mean clustering is a family
characteristic across the OPT scale range, not a property of the 6.7B
model.

\textbf{§5.4's synergy result splits by model and we cannot say why.}
Ablating every GEM together suppresses less than the best single GEM on
both MHA models tested (0/15 and 0/11 atlases positive) and more on the
one GQA model (13/16). Three models --- one of them GQA --- cannot
establish whether that tracks architecture, and this paper has already
retracted one architecture-cohort claim made on a larger sample (§5.1),
so we do not offer it as one. The negative-synergy side is additionally
confounded by patch interference: an upstream ablation shifts the
representation a downstream GEM's stored direction was estimated on, so
that direction under-ablates. The adaptive-direction re-run (§9.4) is
what would separate the two.

\textbf{One concept may be a corpus problem rather than a method
problem.} In the supervised comparison of §8.1, \texttt{deception} is
the only concept on which \emph{both} estimators degrade:
difference-of-means falls to 0.880 and logistic regression to 0.976,
against corpus means of 0.965 and 0.998. It is the only concept where
the supervised probe never reaches ceiling on any of the six models
(0.945--0.991), and the deficit is consistent across architectures and
scales. When a fitted estimator with more free parameters than training
examples cannot separate a concept, the limiting factor is more likely
the labels than the readout. We flag \texttt{deception} as a candidate
for the same pair-quality audit that corrected \texttt{exfiltration}
(§3.5) and do not resolve it here; the concept is retained in all
reported statistics. The other concepts on which the unsupervised
estimator struggles --- threat\_severity, causation, authorization ---
show the opposite signature, with supervised probing at or above its
corpus mean, and are estimator limitations rather than corpus defects
(§8.1).

\textbf{Layer selection does not beat a random same-width window in
every model.} Three of the thirteen models tested fail that null (§5.6)
--- Qwen2.5-3B, Qwen2.5-1.5B and gpt2 --- including Qwen2.5-1.5B, which
clears §5.1's handoff-versus-peak comparison. The aggregate is positive
(169/221 cells, 76.5\%) and the result is stable between a 50-pair pilot
and the 250-pair corpus, so the failures are a scope condition rather
than sampling noise, and we have not established what distinguishes the
models that fail. The weaker limitation is per-pair rather than
per-model: only 18.6\% of cells reach \(p<0.05\) against their own null,
so the aggregate effect is better supported than any individual pair.

\textbf{Scope of ablation claims.} The 66.3\% mean separation reduction
in §5.3 is not the fraction of concept information residing at the
handoff layer. It is the fraction \emph{accessible} from a single
intervention at the most effective site GEM identifies. The remainder is
distributed across the residual stream at layers that contribute
continuously and at levels below the detection threshold of
allocation-event methodology. We would not be surprised if every layer
makes some contribution to every concept --- what CAZ and GEM identify
are the layers where that contribution is large enough to be isolated,
directionally organized, and causally testable. Any single-layer
intervention, of any type, operates on a slice of a continuous process.
That the deepest GEM achieves mean 66.3\% suppression from one slice ---
and that cross-disruption from shallower GEMs reaches the deepest GEM in
37 of 42 multi-GEM atlases across three models (§5.4) --- is what makes
the identified events scientifically useful. It is not a complete
account of concept encoding, and we do not claim it to be.

\textbf{Layer selection is not circular.} GEM selects the handoff layer
where the direction has stopped changing --- mechanically, where it is
best estimated --- and then evaluates that same direction, which admits
a deflationary reading: the advantage might be selection on noise rather
than a claim about where concepts assemble. We tested this by splitting
each concept's pairs in half, detecting the layer on half A, and
extracting, ablating and measuring entirely on half B, so both arms
estimate from B and only the \emph{provenance of the layer index}
differs. Across 134 pairs and 8 models the handoff advantage is 113/134
(84.3\%) with in-sample selection and \textbf{112/134 (83.6\%)
out-of-sample --- a change of −0.7 pp}. The roster for this control is
not listed here; its in-sample base rate (84.3\%) sits well above §5.1's
corpus-wide 69.2\%, and we have not ruled out that the 8 models happen
to be a favorable subset --- the comparison is internal (in-sample
vs.~out-of-sample on the same 8), so it does not depend on that base
rate matching the corpus, but the gap is worth a reader's attention.
Notably, the layer index itself is unstable: the same handoff layer is
chosen from both halves in only 55.2\% of pairs. The result is therefore
not carried by identifying one particular layer, but by locating a
region within which the direction has settled; which layer inside it is
sampled matters little. This is consistent with §5.4's finding that no
single site accounts for the whole signal, and inconsistent with the
selection-on-noise reading.

\textbf{Statistical independence.} The 493 trials in §5.1 are not
exchangeable --- 17 concepts per model produce correlated outcomes, as
the per-model improvement rates demonstrate directly (Llama-3.1-8B:
17/17 = 100\%; pythia-1.4b: 17/17 = 100\%; Qwen2.5-14B: 6/17 = 35\%). A
Wilcoxon signed-rank test on per-model handoff-preference proportions
gives W=356, N=29, p=1.34×10⁻³ (one-sided) --- significant at the model
level. The 23-model corpus achieved significance where the earlier
16-model corpus did not (W=89, p=0.15); the 29-model corpus strengthens
this further. There is no MHA/GQA cohort split at this corpus size:
13/17 MHA models prefer handoff versus 6/9 GQA models (Fisher's exact,
one-sided p=0.462) --- statistically indistinguishable. The published
23-model cohort split (11/13 MHA vs.~2/7 GQA, p=0.022) did not survive
the expanded corpus, and the reason is worth stating precisely, because
it is not simply that the corpus got bigger. Both cohorts reconcile
exactly to the current counts: MHA goes 11/13 → 10/13 on the original
thirteen models plus 3 of the 4 added; GQA goes 2/7 → 4/7 on the
original seven plus 2 of the 2 added. Corpus growth alone --- freezing
the original twenty at their published values --- would give 14/17
vs.~4/9 (p = 0.063), which removes significance but leaves GQA a
minority cohort. The move from there to statistical indistinguishability
comes from \textbf{three of the original twenty models changing verdict
on re-measurement} (one MHA down, two GQA up). Most of the corpus is
stable across vintages --- gpt2-large, gpt2-xl, pythia-1.4b, pythia-12b,
Llama-3.1-8B and opt-6.7b all reproduce their published per-model rates
exactly --- and the exfiltration correction is not the cause, since the
GQA count is 6/9 with or without the exfiltration row. The one
identifiable mover is pythia-160m, which fell from the 59\% reported in
v1 to 29\% at the same N=250 and also accounts for most of the drift in
the sub-500M bin. Because the earlier ablation vintage was overwritten
in place when the corpus was re-run at uniform ablation width, we cannot
attribute that change post hoc; we report it as an unexplained
instability in a small-model row rather than as a corrected error. The
cohort count is in any case fragile at this size: Qwen2.5-1.5B and
Qwen2.5-7B both stand at 9/17, one concept above a strict majority, so a
single concept moves the GQA tally between 6/9 and 4/9. Nor is the
replacement a scale floor: the sub-500M band averages 47\% but is
bimodal, and the corpus-wide scale correlation is weak and non-monotone
(§5.1). We make no scale claim, and the earlier ``\textasciitilde47\%
below 500M, plateauing above'' framing is withdrawn. Handoff depth
distributions shift substantially with N=250 data: most concepts now
have \(L_H/N \geq 0.9\) for the majority of models under the
dominant-GEM convention of §4.3/Table D2, reflecting both later saddle
points in a more stably estimated \(S(l)\) and the structural fact that
each concept's deepest region ends at \(N-1\) by construction (§2.2,
§4.3). This does not affect the ablation results, which are computed at
the actual handoff layer regardless of its depth.

\textbf{Re-run variance is not characterized.} Three of the twenty
original-cohort models changed handoff-preference verdict between the
published corpus and this one's re-measurement at the same N=250, for
reasons that are not reconstructible after the fact (above; pythia-160m
alone moved from 59\% to 29\%). We have no estimate of how often a given
model's per-model tally would change on a repeated run under identical
settings, so individual per-model rates in this paper --- including the
ones used to argue against a scale or cohort effect --- should be read
with that unquantified variance in mind, not as fixed properties of the
model.

\subsubsection{9.4 Open Questions}\label{open-questions}

\begin{itemize}
\tightlist
\item
  Does the near-final rule reflect a genuine model-boundary phenomenon
  or a measurement artifact of the unembedding layer's influence on
  residual streams?
\item
  What is the contribution of attention vs.~MLP sublayers to the handoff
  event? GEM detects it at the residual stream level but does not
  decompose it.
\item
  Do instruct-tuned models show different handoff depths or rotation
  profiles for the same concept?
\item
  \textbf{Adaptive-direction rerun (§5.4):} Re-estimating the settled
  direction at each downstream GEM \emph{after} applying upstream
  patches would resolve whether the apparent anti-synergy is entirely a
  patch-interference artifact or whether some genuine redundancy exists.
  This is the key experiment for establishing whether the multi-GEM
  structure is cooperative or redundant.
\item
  \textbf{Cross-architecture propagation:} Does the GQA pilot's less
  complete cascade propagation (43\% of downstream GEMs vs.~100\% in the
  MHA pilots; Henry, 2026c §6.5) manifest as systematically lower
  cross-disruption in the permutation ablation? §5.4 now measures
  cross-disruption on three models and the answer does not line up
  simply: the one GQA model (Qwen2.5-7B) shows cross-disruption in 16/16
  atlases at the \emph{highest} mean magnitude of the three (0.40), yet
  it is also the only model where ablating all GEMs together suppresses
  more than the best single site. Whether the relay topology is
  architecture-dependent therefore remains open, and needs more than one
  GQA model to settle.
\item
  \textbf{Coverage:} What fraction of total concept signal lies outside
  detectable GEMs? A lower bound could be estimated by comparing the
  separation suppression achieved at the deepest GEM against the
  theoretical maximum from full-layer projection onto the concept
  subspace.
\item
  \textbf{Is the single settled direction the right steering target?} A
  preliminary external steering pilot (sentiment, Qwen2.5-7B-Instruct,
  one model/one concept, not part of this paper's validation) found that
  the GEM settled direction is the stronger single-point intervention in
  the shallow and mid-depth CAZ, but in the deep CAZ a single static
  direction produces almost no steering effect while \emph{distributed}
  per-layer directions tracking the concept's rotation through that zone
  produce the strongest effect observed. If this replicates, it would
  mean the settled-direction abstraction --- while validated here
  geometrically and causally via ablation --- is not uniformly the right
  target for behavioral steering specifically, particularly in CAZs
  where rotation continues later or more steeply. This is an open
  question, not a finding of this paper; response coherence at the
  strongest setting was not independently confirmed in that pilot.
\end{itemize}

\begin{center}\rule{0.5\linewidth}{0.5pt}\end{center}

\subsection{10. Conclusion}\label{conclusion}

Geometric Evolution Maps record a concept's directional trajectory
through assembly rather than sampling it at a single layer. Concept
directions rotate substantially during that assembly (mean entry-exit
cosine 0.285) and only stabilize afterward, so probing during rotation
--- at the separation peak, or at any single layer inside the model ---
is not a reliable way to recover a concept's encoding.

\textbf{GEM's extracted direction is causal, not an artifact of
projection.} Ablating it removes a mean 66.3\% of separation, against
0.07\% for random directions, beating all ten random seeds in every pair
tested (§5.3). That specificity is what validates the method: GEM finds
a real, direction-specific signal, not noise that projecting out any
direction would suppress regardless.

\textbf{No single site is sufficient for that causal effect on its own
--- not the peak, not the handoff, not any other fixed coordinate.} The
handoff-layer probe beats the peak-layer probe in 341/493 trials
(69.2\%; §5.1), but what the corpus supports is that probing
post-settling beats probing during rotation, not that the specific
settling boundary matters beyond the depth it sits at (§5.5) --- a
depth-matched, non-settling control performs indistinguishably from the
handoff layer. Within a concept's atlas the same pattern holds at a
different scale: the deepest GEM is typically the most effective single
ablation site (40/42 atlases, §5.4), but ablating it alone never
approaches complete suppression, and ablating shallower GEMs measurably
changes what remains recoverable at the deepest one (37/42 atlases).
Concept information is distributed across the network; GEM identifies
where that distribution concentrates enough to be isolated, not its
entirety.

\textbf{The consequence for downstream use is that a concept's atlas ---
the full set of its GEMs, not any single one chosen from it --- is the
unit that should carry forward.} This paper delivers that record: each
region's window, its per-layer direction sequence, its settled
direction, and its handoff layer (§3.1). Treating one layer as
sufficient, however it is chosen, discards structure the data says is
there.

GEM provides the ablation-targeting methodology used by the CAZ
Validation paper in this program. Cross-paper relationships are
described in §1.3.

\begin{center}\rule{0.5\linewidth}{0.5pt}\end{center}

\subsection{Appendix A: Model Details}\label{appendix-a-model-details}

\textbf{Primary ablation sub-corpus (§5.2 --- 23 models × 17 concepts =
391 pairs)}

\begin{longtable}[]{@{}
  >{\raggedright\arraybackslash}p{(\columnwidth - 12\tabcolsep) * \real{0.2658}}
  >{\raggedright\arraybackslash}p{(\columnwidth - 12\tabcolsep) * \real{0.1139}}
  >{\raggedright\arraybackslash}p{(\columnwidth - 12\tabcolsep) * \real{0.1013}}
  >{\raggedright\arraybackslash}p{(\columnwidth - 12\tabcolsep) * \real{0.0759}}
  >{\raggedright\arraybackslash}p{(\columnwidth - 12\tabcolsep) * \real{0.1013}}
  >{\raggedright\arraybackslash}p{(\columnwidth - 12\tabcolsep) * \real{0.1899}}
  >{\raggedright\arraybackslash}p{(\columnwidth - 12\tabcolsep) * \real{0.1519}}@{}}
\toprule\noalign{}
\begin{minipage}[b]{\linewidth}\raggedright
Model
\end{minipage} & \begin{minipage}[b]{\linewidth}\raggedright
Family
\end{minipage} & \begin{minipage}[b]{\linewidth}\raggedright
Params
\end{minipage} & \begin{minipage}[b]{\linewidth}\raggedright
N layers
\end{minipage} & \begin{minipage}[b]{\linewidth}\raggedright
Hidden dim
\end{minipage} & \begin{minipage}[b]{\linewidth}\raggedright
Architecture
\end{minipage} & \begin{minipage}[b]{\linewidth}\raggedright
Source
\end{minipage} \\
\midrule\noalign{}
\endhead
\bottomrule\noalign{}
\endlastfoot
pythia-70m & Pythia & 70M & 6 & 512 & MHA & EleutherAI \\
pythia-160m & Pythia & 160M & 12 & 768 & MHA & EleutherAI \\
pythia-410m & Pythia & 410M & 24 & 1024 & MHA & EleutherAI \\
pythia-1b & Pythia & 1B & 16 & 2048 & MHA & EleutherAI \\
pythia-1.4b & Pythia & 1.4B & 24 & 2048 & MHA & EleutherAI \\
pythia-2.8b & Pythia & 2.8B & 32 & 2560 & MHA & EleutherAI \\
pythia-6.9b & Pythia & 6.9B & 32 & 4096 & MHA & EleutherAI \\
pythia-12b & Pythia & 12B & 36 & 5120 & MHA & EleutherAI \\
gpt2 & GPT-2 & 124M & 12 & 768 & MHA & OpenAI \\
gpt2-large & GPT-2 & 774M & 36 & 1280 & MHA & OpenAI \\
gpt2-xl & GPT-2 & 1.5B & 48 & 1600 & MHA & OpenAI \\
opt-1.3b & OPT & 1.3B & 24 & 2048 & MHA & Meta \\
opt-6.7b & OPT & 6.7B & 32 & 4096 & MHA & Meta \\
Qwen2.5-0.5B & Qwen2.5 & 0.5B & 24 & 896 & GQA & Qwen \\
Qwen2.5-1.5B & Qwen2.5 & 1.5B & 28 & 1536 & GQA & Qwen \\
Qwen2.5-3B & Qwen2.5 & 3B & 36 & 2048 & GQA & Qwen \\
Qwen2.5-7B & Qwen2.5 & 7B & 28 & 3584 & GQA & Qwen \\
Qwen2.5-14B & Qwen2.5 & 14B & 48 & 5120 & GQA & Qwen \\
Mistral-7B-v0.3 & Mistral & 7B & 32 & 4096 & GQA & Mistral \\
Llama-3.1-8B & Llama & 8B & 32 & 4096 & GQA & Meta \\
gemma-2-2b & Gemma-2 & 2B & 26 & 2304 & Alternating & Google \\
gemma-2-9b & Gemma-2 & 9B & 42 & 3584 & Alternating & Google \\
phi-2 & Phi & 2.7B & 32 & 2560 & Other & Microsoft \\
\end{longtable}

§5.2's adaptive-width experiment uses the 23-model primary sub-corpus
above (391 pairs), and §5.6's random-window null uses thirteen models in
two batches: 30-window (pythia-1.4b, Llama-3.2-3B, gemma-2-2b,
Qwen2.5-3B, Qwen2.5-1.5B, gpt2) and 50-window (pythia-160m, pythia-70m,
phi-2, pythia-410m, pythia-1b, gpt2-large, opt-1.3b). §5.1's headline
comparison and the §5.3/§5.5 controls use the full 29-model corpus (493
pairs): the 23 models above plus gpt2-medium, opt-125m, opt-350m,
opt-2.7b, Llama-3.2-1B, and Llama-3.2-3B (per-model geometry in Table
D1). Four further large models were excluded for VRAM reasons during
model load (Qwen2.5-32B, Qwen2.5-72B, Llama-3.1-70B, falcon-40b) and
gemma-4-26B-A4B by scope, as a mixture-of-experts configuration (§5.5,
§9.3).

\begin{center}\rule{0.5\linewidth}{0.5pt}\end{center}

\subsection{Appendix B: Concept
Inventory}\label{appendix-b-concept-inventory}

The 17 concepts in the current corpus span semantic, epistemic,
pragmatic, and safety domains.

\begin{longtable}[]{@{}
  >{\raggedright\arraybackslash}p{(\columnwidth - 4\tabcolsep) * \real{0.3214}}
  >{\raggedright\arraybackslash}p{(\columnwidth - 4\tabcolsep) * \real{0.2143}}
  >{\raggedright\arraybackslash}p{(\columnwidth - 4\tabcolsep) * \real{0.4643}}@{}}
\toprule\noalign{}
\begin{minipage}[b]{\linewidth}\raggedright
Concept
\end{minipage} & \begin{minipage}[b]{\linewidth}\raggedright
Type
\end{minipage} & \begin{minipage}[b]{\linewidth}\raggedright
Description
\end{minipage} \\
\midrule\noalign{}
\endhead
\bottomrule\noalign{}
\endlastfoot
agency & Semantic & Attribution of intentional action to an agent \\
authorization & Semantic & Framing of permission or legitimate
authority \\
causation & Relational & Causal dependency between events \\
certainty & Epistemic & Degree of confidence or assertion strength \\
credibility & Epistemic & Reliability or trustworthiness of a source \\
deception & Pragmatic & Intentional misleading framing \\
exfiltration & Safety & Context of unauthorized data or information
transfer \\
formality & Pragmatic & Register level of discourse \\
moral\_valence & Affective & Ethical evaluation (positive/negative) \\
negation & Syntactic & Logical or linguistic negation \\
plurality & Syntactic & Singular vs.~plural reference \\
sarcasm & Pragmatic & Ironic or non-literal intent \\
sentiment & Affective & Emotional tone (positive/negative) \\
specificity & Pragmatic & Level of detail or precision \\
temporal\_order & Relational & Sequencing of events in time \\
threat\_severity & Safety & Severity or imminence of a threat \\
urgency & Safety & Time-pressure or urgency framing \\
\end{longtable}

\begin{center}\rule{0.5\linewidth}{0.5pt}\end{center}

\subsection{Appendix C: Contrastive Pair
Construction}\label{appendix-c-contrastive-pair-construction}

Pairs are drawn from the
\href{https://github.com/jamesrahenry/Rosetta_Concept_Pairs}{Rosetta
Concept Pairs dataset} (DOI:
\href{https://doi.org/10.5281/zenodo.20059650}{10.5281/zenodo.20059650}).
Each pair consists of a positive sentence (strongly expressing the
target concept) and a negative sentence (same topic, concept absent),
constructed via a multi-model consensus protocol: for each concept and
topic, multiple large language models from the Claude, GPT, Gemini,
Kimi, and Mistral families independently produced candidate pairs, and
pairs were retained only when all contributing models agreed on the
concept label (unanimous agreement across the generator models that
produced the pair). The generator roster is 14 models spanning five AI
labs --- Anthropic (4), OpenAI (5), Google (3), Mistral (1) and Moonshot
(1) --- and is identical across all 17 concepts; a mean of approximately
12 of the 14 contribute to any given topic (per-concept range 8--13), as
recorded in the Rosetta Concept Pairs dataset metadata. This cross-model
consistency filter is the operational ambiguity exclusion criterion ---
pairs where any model assigned a different label were discarded.

Passages are paragraph-length (approximately 150--350 tokens). Each pair
shares the same topic and surface register, differing specifically along
the target concept dimension. Each concept in this paper draws 250
consensus-validated pairs, with the exception of exfiltration at 249
following the label correction (§3.5); the same pair set is used for all
models.

Activations are extracted as the last-token representation from the
post-MLP residual stream output of each transformer block (the full
block output, after both attention and feed-forward sub-layers), using
bfloat16 forward passes and float32 metric computation, via the shared
extraction tooling for this research program {[}Henry, 2026,
software{]}. Models are set to inference mode (no dropout). Passages are
tokenized using each model's default tokenizer with default BOS token
handling; no system prompt, instruction prefix, or task framing is
prepended. For models distributed across multiple GPUs under
\texttt{device\_map="auto"}, cross-device tensor operations are handled
explicitly.

\textbf{Layer indexing.} Layer indices in this paper run
\(l \in \{0, \dots, N-1\}\) over transformer \emph{blocks}, where \(N\)
is the model's \texttt{num\_hidden\_layers}. The embedding output is
\textbf{not} a layer: a forward pass exposing \texttt{hidden\_states}
returns \(N+1\) tensors, of which the first (the embedding) is dropped
before any index in this paper is assigned. A reimplementation that
indexes \texttt{hidden\_states} directly will be off by one at every
layer, including \(L_{\text{peak}}\), \(L_{\text{CAZ\_end}}\) and
\(L_H\).

\textbf{Tokenization limits.} Passages are truncated at
\texttt{max\_length=512} tokens. Given the 150--350 token passage length
above this is rarely binding, but it is not never-binding, and the
last-token representation is taken at the last \emph{real} token --- the
final non-padding position --- rather than at the last position in the
padded tensor.

\begin{center}\rule{0.5\linewidth}{0.5pt}\end{center}

\subsection{Data and Code
Availability}\label{data-and-code-availability}

The reusable extraction and analysis library is \texttt{rosetta\_tools}
v1.3.1 (DOI: https://doi.org/10.5281/zenodo.20361433); the analysis,
figure, and reproduction scripts are \texttt{Rosetta\_Analysis} v1.1.0
(DOI: https://doi.org/10.5281/zenodo.21583139), which imports that
library; and the umbrella index for the series is at
https://github.com/jamesrahenry/Rosetta. Pre-extracted model activations
for all 29 corpus models across 17 concepts are available as the Rosetta
Activations dataset at
https://huggingface.co/datasets/james-ra-henry/Rosetta-Activations (DOI:
https://doi.org/10.57967/hf/9725). Contrastive pairs are available at
https://github.com/jamesrahenry/Rosetta\_Concept\_Pairs (DOI:
https://doi.org/10.5281/zenodo.20059650).

\subsection{Acknowledgments}\label{acknowledgments}

The author acknowledges the support of TELUS, specifically the Chief AI
Office, the AI Accelerator, and the Chief Security Office.

Thanks to Ivey Chiu, Steve Pearson, and Krista Hickey for helpful
discussions and guidance.

The author acknowledges computational and academic support from the
Vector Institute for Artificial Intelligence.

Claude (Anthropic) contributed substantially to this work, including
analysis design, validation methodology, and manuscript development.

\textbf{Competing interests.} The author is an employee of TELUS
Communications Inc.; this research was conducted independently of that
role, and the author declares no competing interests arising from it.
TELUS and the Vector Institute are acknowledged for their support (see
Acknowledgments); neither had any role in the study's design, data
collection, analysis, interpretation, or the decision to publish.

\begin{center}\rule{0.5\linewidth}{0.5pt}\end{center}

\subsection{Appendix D: Per-Model and Per-Concept
Results}\label{appendix-d-per-model-and-per-concept-results}

Evidence tables supporting §4.1--§4.3 (Tables D1--D2) and §5.3--§5.6
(Tables D3--D6). Every figure cited in the main text appears here in
full; the main text carries one representative number per claim and
points to this appendix for the distribution behind it.

\subsubsection{D.1 Per-model rotation}\label{d.1-per-model-rotation}

\textbf{Table D1: Mean entry-exit cosine per model (lower = more
rotation), with per-layer rotation rate.} Supports §4.2.

\begin{longtable}[]{@{}llll@{}}
\toprule\noalign{}
Model & Params & Mean EEC & Rotation/layer \\
\midrule\noalign{}
\endhead
\bottomrule\noalign{}
\endlastfoot
gpt2 & 124M & 0.103 & 0.586 \\
gpt2-medium & 355M & 0.104 & 0.502 \\
Llama-3.1-8B & 8B & 0.205 & 0.343 \\
Llama-3.2-1B & 1B & 0.216 & 0.442 \\
pythia-2.8b & 2.8B & 0.224 & 0.336 \\
pythia-1.4b & 1.4B & 0.229 & 0.311 \\
gpt2-xl & 1.5B & 0.236 & 0.307 \\
opt-2.7b & 2.7B & 0.236 & 0.288 \\
pythia-12b & 12B & 0.237 & 0.257 \\
pythia-6.9b & 6.9B & 0.239 & 0.283 \\
pythia-70m & 70M & 0.249 & 0.541 \\
Llama-3.2-3B & 3B & 0.255 & 0.305 \\
opt-6.7b & 6.7B & 0.258 & 0.259 \\
opt-1.3b & 1.3B & 0.263 & 0.273 \\
pythia-1b & 1B & 0.265 & 0.362 \\
opt-125m & 125M & 0.272 & 0.189 \\
opt-350m & 350M & 0.280 & 0.155 \\
phi-2 & 2.7B & 0.290 & 0.166 \\
pythia-410m & 410M & 0.308 & 0.283 \\
Mistral-7B-v0.3 & 7B & 0.319 & 0.222 \\
gpt2-large & 774M & 0.325 & 0.211 \\
Qwen2.5-14B & 14B & 0.326 & 0.186 \\
gemma-2-9b & 9B & 0.348 & 0.118 \\
Qwen2.5-7B & 7B & 0.351 & 0.312 \\
gemma-2-2b & 2B & 0.372 & 0.172 \\
Qwen2.5-0.5B & 0.5B & 0.403 & 0.331 \\
pythia-160m & 160M & 0.421 & 0.396 \\
Qwen2.5-1.5B & 1.5B & 0.441 & 0.246 \\
Qwen2.5-3B & 3B & 0.489 & 0.185 \\
\end{longtable}

\subsubsection{D.2 Per-concept handoff
depth}\label{d.2-per-concept-handoff-depth}

\textbf{Table D2: Mean handoff depth by concept (23 models).} Supports
§4.3; the shallow/mid/deep grouping quoted there is derived from the
mean column.

\begin{longtable}[]{@{}lll@{}}
\toprule\noalign{}
Concept & Mean rel. depth & Median rel. depth \\
\midrule\noalign{}
\endhead
\bottomrule\noalign{}
\endlastfoot
specificity & 0.629 & 0.607 \\
plurality & 0.717 & 0.750 \\
sentiment & 0.717 & 0.833 \\
moral\_valence & 0.728 & 0.917 \\
formality & 0.731 & 0.750 \\
credibility & 0.739 & 0.750 \\
deception & 0.782 & 0.958 \\
sarcasm & 0.786 & 0.917 \\
urgency & 0.796 & 0.958 \\
agency & 0.801 & 0.938 \\
negation & 0.814 & 0.958 \\
exfiltration & 0.825 & 0.917 \\
causation & 0.841 & 0.964 \\
authorization & 0.865 & 0.964 \\
temporal\_order & 0.874 & 0.962 \\
certainty & 0.895 & 0.964 \\
threat\_severity & 0.911 & 0.962 \\
\end{longtable}

\textbf{Pythia scale-ladder depth, by node convention (supports §6.1).}
Under the shallowest-GEM convention (§5.2's
\texttt{ablation\_targets{[}0{]}}), mean relative handoff depth across
the ladder (70M → 160M → 410M → 1B → 1.4B → 2.8B → 6.9B → 12B) runs
0.627, 0.392, 0.424, 0.588, 0.559, 0.467, 0.331, 0.412 --- a fall of
0.216 from smallest to largest. Averaging over every GEM in each atlas
gives the same direction (0.730 → 0.681, endpoints only). Two other
conventions rise on net instead: dominant-GEM (0.725 → 0.874) and
deepest-GEM (0.833 → 0.972), though neither monotonically, and the
deepest-GEM series is dominated by the construction artifact of §3.2,
since every atlas's deepest segment terminates at \(N-1\) by definition.
Which direction the apparent ``trend'' runs is therefore a choice of
convention, not a property of the models.

The near-final trigger count shows the same non-ordering: it falls from
6 concepts at pythia-1b to 5 at pythia-2.8b (the near-final rule is
already firing at 1B), and reaches 7 at pythia-12b --- it does not order
by scale either.

\subsubsection{D.3 Direction specificity by architecture
cohort}\label{d.3-direction-specificity-by-architecture-cohort}

\textbf{Table D3: Direction-specificity control by cohort (handoff
layer, C=17, n=472).} Supports §5.3. The \emph{median ratio} column is
not comparable across cohorts --- cohorts differ in mean hidden width
and the ratio tracks width (§5.3); the concept, random and
\textgreater all-seeds columns are the interpretable ones.

\begin{longtable}[]{@{}lllllll@{}}
\toprule\noalign{}
Cohort & N & Concept & Random & Median ratio & Median z &
\textgreater all seeds \\
\midrule\noalign{}
\endhead
\bottomrule\noalign{}
\endlastfoot
MHA & 268 & 70.7\% & 0.07\% & 2,370.7× & 1,231.7 & 100\% \\
GQA & 153 & 64.3\% & 0.08\% & 3,175.9× & 1,771.3 & 100\% \\
Gemma & 34 & 38.6\% & 0.05\% & 1,249.9× & 619.2 & 100\% \\
Phi-2 & 17 & 71.0\% & 0.07\% & 1,451.8× & 834.0 & 100\% \\
\end{longtable}

\subsubsection{D.4 Multi-GEM ablation: corpus
detail}\label{d.4-multi-gem-ablation-corpus-detail}

Supporting detail for §5.4.

\textbf{Provenance.} The §5.4 figures come from a permutation ablation
re-run at the corpus standard of \textbf{250 pairs per concept}
(2026-07-29), superseding an earlier 50-pair run on pythia-6.9b alone.
The generating scripts:

\begin{itemize}
\tightlist
\item
  GEM counts: \texttt{gem/regenerate\_table3\_handoff\_depth.py}
\item
  Permutation results: \texttt{gem/ablate\_gem\_permutation.py}, driven
  by the flow script \texttt{flow\_p2\_permutation\_refreeze.py} (in
  \texttt{validation/p2\_gem/regeneration/})
\end{itemize}

\textbf{GEM-count distribution (pythia-6.9b, 17 concepts).} 2 GEMs in 13
concepts, 3 in 2 (negation, sentiment), 1 in 2 (authorization,
threat\_severity) --- 15 multi-GEM atlases. gpt2-xl contributes 11 and
Qwen2.5-7B 16, for the 42 atlases §5.4 reports.

\textbf{Table D4: Multi-GEM ablation, per-model results (250 pairs).}
Supports §5.4.

\begin{longtable}[]{@{}
  >{\raggedright\arraybackslash}p{(\columnwidth - 14\tabcolsep) * \real{0.1250}}
  >{\raggedright\arraybackslash}p{(\columnwidth - 14\tabcolsep) * \real{0.1250}}
  >{\raggedright\arraybackslash}p{(\columnwidth - 14\tabcolsep) * \real{0.1250}}
  >{\raggedright\arraybackslash}p{(\columnwidth - 14\tabcolsep) * \real{0.1250}}
  >{\raggedright\arraybackslash}p{(\columnwidth - 14\tabcolsep) * \real{0.1250}}
  >{\raggedright\arraybackslash}p{(\columnwidth - 14\tabcolsep) * \real{0.1250}}
  >{\raggedright\arraybackslash}p{(\columnwidth - 14\tabcolsep) * \real{0.1250}}
  >{\raggedright\arraybackslash}p{(\columnwidth - 14\tabcolsep) * \real{0.1250}}@{}}
\toprule\noalign{}
\begin{minipage}[b]{\linewidth}\raggedright
Model
\end{minipage} & \begin{minipage}[b]{\linewidth}\raggedright
Attention
\end{minipage} & \begin{minipage}[b]{\linewidth}\raggedright
Layers
\end{minipage} & \begin{minipage}[b]{\linewidth}\raggedright
Multi-GEM atlases
\end{minipage} & \begin{minipage}[b]{\linewidth}\raggedright
Deepest dominates
\end{minipage} & \begin{minipage}[b]{\linewidth}\raggedright
Cross-disruption
\end{minipage} & \begin{minipage}[b]{\linewidth}\raggedright
Mean synergy
\end{minipage} & \begin{minipage}[b]{\linewidth}\raggedright
Positive
\end{minipage} \\
\midrule\noalign{}
\endhead
\bottomrule\noalign{}
\endlastfoot
pythia-6.9b & MHA & 32 & 15 & 15/15 & 11/15 (0.253) & −0.103 & 0/15 \\
gpt2-xl & MHA & 48 & 11 & 11/11 & 10/11 (0.15) & −0.089 & 0/11 \\
Qwen2.5-7B & GQA & 28 & 16 & 14/16 & 16/16 (0.40) & +0.061 & 13/16 \\
\end{longtable}

The synergy sign splits with attention type across these three models.
\textbf{§5.4 declines to read that as a cohort effect and so does this
appendix}: it is one GQA model against two MHA models, and §5.1
retracted an architecture-cohort claim made on a substantially larger
sample. An earlier version of this appendix offered a mechanism for the
split --- grouped key-value sharing concentrating concept routing
through fewer attention paths, so that disruption propagates rather than
being absorbed. \textbf{That explanation is withdrawn.} It is untested,
it was constructed after seeing the sign, and three models cannot
support it.

\textbf{Companion context.} A parallel dependency analysis in 28 base
models {[}Henry, 2026c §6.2{]} finds forward dependency to be a minority
phenomenon --- 17.6\% of the 850 testable CAZ pairs (equivalently 8.8\%
of all 1,700 directed pairs), 82.4\% independent, 0\%
backward-dependent. In that paper's cascade-propagation pilots (§6.5),
ablating the upstream handoff propagated to 100\% of dependent
downstream GEMs in two MHA pilot models (pythia-1.4b, gpt2-xl) and 43\%
in one GQA pilot (Qwen2.5). Our GQA model shows \emph{stronger}
per-atlas cross-disruption than either MHA model, which does not
obviously sit with that 43\%. We record the tension and do not resolve
it; both observations rest on single-model pilots.

\subsubsection{D.5 Depth-matched comparison: per-model
detail}\label{d.5-depth-matched-comparison-per-model-detail}

\textbf{Table D5: Depth-matched comparison, per-model win rate.}
Supporting detail for §5.5, under the \textbf{site-matched} control (one
depth-matched control layer per GEM, 17 concepts per model, 29 models,
493 pairs). Per-model win rate against the control and mean advantage:

\begin{longtable}[]{@{}llllll@{}}
\toprule\noalign{}
Model & Wins & Mean Δ & Model & Wins & Mean Δ \\
\midrule\noalign{}
\endhead
\bottomrule\noalign{}
\endlastfoot
pythia-1b & 17/17 & +20.4pp & opt-1.3b & 11/17 & +3.9pp \\
opt-125m & 17/17 & +19.0pp & pythia-1.4b & 11/17 & +3.2pp \\
Llama-3.1-8B & 17/17 & +11.4pp & opt-6.7b & 11/17 & +0.2pp \\
Llama-3.2-1B & 16/17 & +16.1pp & opt-2.7b & 11/17 & +0.1pp \\
Llama-3.2-3B & 15/17 & +7.5pp & pythia-12b & 9/17 & +1.2pp \\
pythia-2.8b & 15/17 & +4.9pp & pythia-160m & 9/17 & +0.6pp \\
pythia-410m & 14/17 & +4.8pp & pythia-70m & 8/17 & +7.9pp \\
pythia-6.9b & 13/17 & +3.5pp & opt-350m & 8/17 & +0.3pp \\
Qwen2.5-7B & 12/17 & +8.3pp & gpt2-large & 6/17 & −1.9pp \\
Qwen2.5-0.5B & 12/17 & +0.1pp & Qwen2.5-14B & 6/17 & −4.2pp \\
Qwen2.5-1.5B & 6/17 & −4.3pp & Qwen2.5-3B & 5/17 & −2.4pp \\
gpt2-xl & 5/17 & −3.7pp & Mistral-7B-v0.3 & 4/17 & −3.0pp \\
phi-2 & 2/17 & −4.0pp & gemma-2-9b & 1/17 & −16.9pp \\
gpt2 & 1/17 & −73.1pp & gpt2-medium & 1/17 & −145.0pp \\
gemma-2-2b & 0/17 & −20.9pp & & & \\
\end{longtable}

Four models carry documented measurement problems that make their cells
uninformative here rather than negative results: gpt2 and gpt2-medium
show the §7.3 ablation pathology, which is what produces their very
large negative means, and the two Gemma-2 models carry the §7.4
calibration limit. Across the remaining 25 models (excluding gpt2,
gpt2-medium and both Gemma-2 models) the rate is 260/425 (61.2\%) at a
mean of +3.6pp --- see below for why that figure is not robust to the
choice of roster or aggregator. The pattern does not track architecture
family or scale: the three models at 100\% span Pythia, OPT, and Llama,
and 125M--8B parameters, while Qwen2.5 sits below 50\% at three of five
scales (1.5B, 3B, 14B; the 0.5B and 7B rows clear a majority).

\textbf{Roster and aggregator robustness (supporting §5.5).} Two
exclusion rosters are defensible and both are pre-existing rather than
chosen for this check: this paper's, dropping gpt2 and gpt2-medium for
the §7.3 pathology and the two Gemma-2 models for the §7.4 calibration
limit; and the companion validation paper's base-model roster, which
instead drops opt-350m for a non-uniform activation-dimension defect,
does not contain Llama-3.1-8B, and excludes both Gemma-2 models for the
same estimator-reliability reason as §7.4 (verified against the
per-model win table above: 263 total wins minus opt-350m (8),
Llama-3.1-8B (17), gemma-2-2b (0) and gemma-2-9b (1) gives exactly the
row's 237). Both happen to be 25 models and 425 pairs, and they share 23
of 25 --- the two Gemma-2 models are this roster's only overlap with
this paper's own exclusion set.

Two per-model aggregators are likewise defensible: the win-rate Wilcoxon
against 0.5, and the median-delta Wilcoxon against 0 that the companion
paper adopts as primary.

\begin{longtable}[]{@{}
  >{\raggedright\arraybackslash}p{(\columnwidth - 8\tabcolsep) * \real{0.2000}}
  >{\raggedright\arraybackslash}p{(\columnwidth - 8\tabcolsep) * \real{0.2000}}
  >{\raggedright\arraybackslash}p{(\columnwidth - 8\tabcolsep) * \real{0.2000}}
  >{\raggedright\arraybackslash}p{(\columnwidth - 8\tabcolsep) * \real{0.2000}}
  >{\raggedright\arraybackslash}p{(\columnwidth - 8\tabcolsep) * \real{0.2000}}@{}}
\toprule\noalign{}
\begin{minipage}[b]{\linewidth}\raggedright
roster
\end{minipage} & \begin{minipage}[b]{\linewidth}\raggedright
wins
\end{minipage} & \begin{minipage}[b]{\linewidth}\raggedright
mean Δ
\end{minipage} & \begin{minipage}[b]{\linewidth}\raggedright
win-rate \(p\)
\end{minipage} & \begin{minipage}[b]{\linewidth}\raggedright
median \(p\)
\end{minipage} \\
\midrule\noalign{}
\endhead
\bottomrule\noalign{}
\endlastfoot
all 29 models & 263/493 (53.3\%) & −5.7pp & 0.456 & 0.804 \\
25, excluding gpt2, gpt2-medium and both Gemma-2 (this paper) & 260/425
(61.2\%) & +3.6pp & 0.037 & 0.104 \\
25, the companion's base roster (excludes opt-350m, Llama-3.1-8B, and
both Gemma-2 models) & 237/425 (55.8\%) & −5.6pp & 0.242 & 0.563 \\
23, excluded by either paper's criteria & 235/391 (60.1\%) & +3.4pp &
0.055 & 0.162 \\
\end{longtable}

No cell above is emphasized, deliberately. The residual clears
\(p < 0.05\) in exactly one of the eight roster × aggregator
combinations --- this paper's roster under this paper's aggregator ---
and in none of the other seven. A single significant cell in a grid
built to test whether the result is stable is evidence that it is not,
and §5.5 does not claim it.

The per-model figures for the superseded single-site control are not
reproduced here; that comparison is withdrawn as a measure of the
settling criterion (§5.5) and only its corpus totals are retained, in
the §5.5 results table.

\subsubsection{D.6 Random-window null: per-model
detail}\label{d.6-random-window-null-per-model-detail}

\textbf{Table D6: GEM handoff versus a random same-width window (250
pairs).} Supports §5.6. The two batches differ only in null size, which
moves the \(p\)-floor (30 windows → 0.032; 50 → 0.020) but not the ratio
or the beats-null count. gpt2's zero ratio is the §7.3 ablation
pathology, not a measured null result.

\emph{30-window batch (6 models, 68/102 cells beat the null):}

\begin{longtable}[]{@{}lllll@{}}
\toprule\noalign{}
Model & Beats null & Median ratio & Median \(p\) & \(p<0.05\) \\
\midrule\noalign{}
\endhead
\bottomrule\noalign{}
\endlastfoot
gemma-2-2b & 16/17 & 2.40 & 0.067 & 8/17 \\
Llama-3.2-3B & 17/17 & 2.10 & 0.133 & 5/17 \\
pythia-1.4b & 17/17 & 1.95 & 0.067 & 6/17 \\
Qwen2.5-3B & 6/17 & 0.91 & 0.633 & 1/17 \\
Qwen2.5-1.5B & 6/17 & 0.85 & 0.567 & 4/17 \\
gpt2 & 6/17 & 0.00 & 1.000 & 0/17 \\
\end{longtable}

\emph{50-window batch (7 models, 101/119 cells beat the null):}

\begin{longtable}[]{@{}lllll@{}}
\toprule\noalign{}
Model & Beats null & Median ratio & Median \(p\) & \(p<0.05\) \\
\midrule\noalign{}
\endhead
\bottomrule\noalign{}
\endlastfoot
pythia-160m & 17/17 & 1.63 & 0.100 & 2/17 \\
pythia-70m & 17/17 & 1.82 & 0.220 & 0/17 \\
phi-2 & 16/17 & 2.07 & 0.120 & 6/17 \\
pythia-410m & 15/17 & 1.76 & 0.180 & 2/17 \\
pythia-1b & 14/17 & 2.36 & 0.100 & 1/17 \\
gpt2-large & 13/17 & 1.54 & 0.260 & 4/17 \\
opt-1.3b & 9/17 & 1.27 & 0.360 & 2/17 \\
\end{longtable}

\subsection{Appendix E: Changes in Version
2}\label{appendix-e-changes-in-version-2}

This version corrects and retracts claims made in v1 (arXiv:2605.25848),
the version currently served by arXiv. A reader holding v1 should treat
the following as superseded. Changes are grouped by \emph{why} the value
moved, because that distinction matters: some numbers changed because
the corpus grew, and others because v1 was wrong.

\textbf{Claims withdrawn.} Each of these was asserted in v1 or in an
interim revision and is no longer supported:

\begin{itemize}
\tightlist
\item
  \textbf{The MHA/GQA cohort split.} v1 reported 11/13 MHA vs.~2/7 GQA
  models preferring the handoff (Fisher one-sided p = 0.022) and treated
  it as a pronounced architectural effect. At 29 models the rates are
  13/17 vs.~6/9 (p = 0.462) --- statistically indistinguishable. §5.1
  decomposes what changed; corpus growth accounts for part of it but not
  all.
\item
  \textbf{The scale account.} v1 stated that ``the residual gradient
  reflects genuine scale effects'' and was ``consistent with a genuine
  scale threshold around 400--500M''; an interim revision promoted this
  to a sub-500M floor once the cohort split was retracted. Both are
  withdrawn: the band is bimodal rather than a floor, three of the seven
  sub-500M models reach 76--88\%, and depth does not explain the low
  cluster (§5.1).
\item
  \textbf{The ε = 0.05 angular-velocity handoff detector.} The handoff
  layer is derived from the CAZ region boundary and never was detected
  by a rotation-threshold criterion; §2.2/§3.2 now state the actual
  construction.
\item
  \textbf{The direction-specificity ratio as an effect size.} v1
  reported a median 377× and read increases in it as evidence about
  concept encoding. The random-direction baseline sits at the \(1/d\)
  chance level, so the ratio scales with hidden width (§5.3). The
  comparison against the pilot value is withdrawn, as is any
  cross-cohort reading of it.
\item
  \textbf{The v1 interpretation of Gemma-2.} v1 stated Gemma had ``the
  highest specificity ratio (863×) \ldots{} consistent with Gemma's more
  geometrically precise concept encoding.'' That reading is
  \textbf{retracted in both directions}: neither it nor its opposite is
  supported, because a single settled direction is not stably estimable
  for this family at N = 250 (§7.4).
\item
  \textbf{The claim that both ablation arms were scored at their own
  extraction layers.} They were scored at the final layer. §5.1 states
  the actual measurement layer and the depth confound it implies.
\item
  \textbf{Superiority over delta-PCA / windowed-PCA probing} (§8.3).
  Never implemented, never measured.
\item
  \textbf{The depth-matched control's confirmation of the settling
  criterion.} v1's Contribution 7 stated that ``a confound test across
  374 concept × model pairs (29 models) confirms that the GEM settling
  criterion contributes +32.5pp mean additional suppression beyond
  probing at a depth-matched non-settling layer.'' That control ablated
  \textbf{one} layer against a treatment arm that ablates every GEM in
  the atlas, and matched depth to the \emph{shallowest} of those sites.
  Re-run with one depth-matched control layer per GEM, the advantage
  falls to +3.6pp on the 25 models without a documented measurement
  problem and is absent on the full corpus (§5.5). The +32.5pp figure
  did not reproduce even under the original design, which gave +41.9pp
  on the current corpus.
\item
  \textbf{OPT's high-EEC architectural profile.} v1 stated that
  ``several concepts genuinely rotate less during assembly than in
  Pythia or Llama-family models'' and that this is ``a real property of
  OPT's architecture and training.'' All five OPT models sit
  \emph{below} the corpus mean EEC of 0.285 --- they rotate more than
  average, not less --- and opt-6.7b is below two of the corpus's three
  Llama models on the same statistic, though not the third, Llama-3.2-3B
  (Table D1). The family-level claim is withdrawn; only a concept-level
  observation survives (§7.2).
\item
  \textbf{The Pythia-ladder depth trend.} v1 reported that within the
  Pythia scale ladder ``mean ablation depth follows an uneven but
  increasing trend,'' with ``the most notable jump'' between 1B and
  2.8B. Under the convention in which \(L_H/N\) is actually evaluated
  the trend runs the other way (0.627 → 0.412 from 70M to 12B), and the
  near-final trigger count \emph{falls} across that ``jump'' (6 at
  pythia-1b, 5 at pythia-2.8b). Withdrawn (§6.1).
\item
  \textbf{``EEC and handoff depth are independent dimensions'' (§9.2).}
  v1 asserted this on the basis of which concepts rotate least, and
  named the wrong ones. The two quantities are in fact strongly
  associated across the 17 concepts (\(\rho = -0.752\), \(p = 0.0005\)),
  and the association survives controlling for segment span
  (\(\rho = -0.540\), \(p = 0.031\)). The independence claim is
  withdrawn; §9.2 now reports the association and declines only the
  causal reading.
\item
  \textbf{The ``SWA exceptions'' reading of §5.5's two depth-matched
  exceptions.} v1 grouped gemma-2-2b and Mistral-7B-v0.3 as sharing
  ``non-standard sliding window attention'' and explained both
  exceptions by it. \textbf{Mistral-7B-v0.3 does not use sliding window
  attention} --- it was removed from the Mistral 7B line at v0.2, and
  v0.3 sets \texttt{sliding\_window:\ null}. The model is standard GQA
  with full attention. The grouping and its explanation are withdrawn;
  gemma-2-2b's exception is re-sourced to the calibration limit of §7.4
  and Mistral-7B-v0.3's is reported as unexplained (§5.5).
\item
  \textbf{gpt2's ablation pathology attributed to depth alone.} v1
  stated ``the root cause is depth: gpt2's 12-layer architecture places
  the handoff in the penultimate layer\ldots{} the near-final rule
  (§5.2) is designed to flag models with N \textless{} 20 layers for
  this reason.'' Neither half holds. gpt2-medium (24 layers) shows the
  same pathology in 10/17 pairs while other 12-layer models in the
  corpus (pythia-160m, opt-125m) show none, so depth alone does not
  explain which models fail --- it tracks the GPT-2 family specifically,
  for a mechanism this paper does not establish (§7.3). And the
  evaluated near-final rule --- the bare \(L_H/N\) \textgreater{} 0.85
  threshold --- has no N \textless{} 20 term; that belongs to a
  separate, recommended-but-unevaluated refinement. The evaluated rule
  \emph{fails} on gpt2 rather than flagging it for exclusion (§5.2).
\item
  \textbf{The settled direction ``remains stable through the remaining
  depth.''} v1's introduction and its per-layer trajectory sketch both
  asserted this (``\emph{settle} into a direction that remains stable
  through the remaining depth''; ``Maintains that direction through the
  remaining depth with slow drift''). This paper measures only the
  single step immediately past the handoff layer (the handoff cosine,
  §3.2): mean 0.828 on the 598 informative GEMs, only 51.7\% exceeding
  0.85. Whether the direction persists further --- to the output, or
  across the full remaining depth --- is not measured here, and the
  stability-through-remaining-depth claim is withdrawn; §1/§2.3 now
  state only that the direction is markedly more stable than during
  assembly, with further persistence left untracked.
\item
  \textbf{Pythia-2.8b's seventeen 1.0 handoff cosines read as evidence
  of an abrupt settling event.} An earlier version read these (17 of its
  33 GEMs, 51.5\%) this way, but \textbf{all seventeen are terminal
  GEMs}, one per concept, and are 1.000 by construction (§3.2) --- the
  comparison is the final layer against itself, so they are evidence of
  nothing. On the sixteen informative GEMs the handoff cosine ranges
  from 0.677 to 0.957 (mean 0.827), which neither supports nor rules out
  an abrupt settling event (§9.1).
\item
  \textbf{Two arguments against the EEC-depth association (§9.2).} An
  earlier version argued the per-layer rotation rate's non-significant
  relationship with depth undercut the association, and that this
  robustness held at the concept level. Neither holds: its sign is
  \textbf{positive}, which for a rotation \emph{rate} corroborates
  rather than undercuts the association (a non-significant result on 17
  points is weak evidence either way), and the robustness premise
  (§4.1's \(r = -0.07\)) is a per-GEM correlation --- it is robust to
  span at the GEM level, not at the concept level, where the same
  quantity correlates with span at \(\rho \approx +0.3\) instead.
  Withdrawn.
\end{itemize}

\textbf{Corrections --- v1 was wrong or under-specified, independent of
corpus size.}

\begin{longtable}[]{@{}
  >{\raggedright\arraybackslash}p{(\columnwidth - 6\tabcolsep) * \real{0.2500}}
  >{\raggedright\arraybackslash}p{(\columnwidth - 6\tabcolsep) * \real{0.2500}}
  >{\raggedright\arraybackslash}p{(\columnwidth - 6\tabcolsep) * \real{0.2500}}
  >{\raggedright\arraybackslash}p{(\columnwidth - 6\tabcolsep) * \real{0.2500}}@{}}
\toprule\noalign{}
\begin{minipage}[b]{\linewidth}\raggedright
quantity
\end{minipage} & \begin{minipage}[b]{\linewidth}\raggedright
v1
\end{minipage} & \begin{minipage}[b]{\linewidth}\raggedright
v2
\end{minipage} & \begin{minipage}[b]{\linewidth}\raggedright
why
\end{minipage} \\
\midrule\noalign{}
\endhead
\bottomrule\noalign{}
\endlastfoot
Entry-exit cosine (mean) & 0.233 & \textbf{0.285} & one aggregation
convention, applied consistently (§4.1) \\
Handoff cosine (mean) & 0.942 & \textbf{0.828} & GEM-level convention;
the terminal-layer degeneracy is now excluded and disclosed (§3.2) \\
Direction-specificity control coverage & ``111 base-model pairs across
16 models'' & \textbf{472 pairs, C = 17, 29 models} & v1's control
covered \textbf{7 of 17 concepts} and did not say so (§5.3) \\
Falsification criterion & stated, then met and explained away &
\textbf{conceded} (§7.5) & gpt2, gpt2-medium and pythia-70m satisfy
it \\
Trial-level Wilcoxon (p = 3.21 × 10⁻¹⁷) & reported & \textbf{dropped} &
pseudoreplication across 17 correlated concepts per model; the
model-level test is the valid one \\
exfiltration pairs & 250 & \textbf{249} & label correction (§3.5) \\
pythia-160m handoff preference & 59\% & \textbf{29\%} & changed on
re-measurement at the same N=250, for reasons not reconstructible after
the corpus was re-run in place (§9.3) \\
\end{longtable}

\textbf{Changes attributable to the larger corpus.} The primary corpus
grew from 23 to 29 base models (391 → 493 concept × model pairs). The
headline moved from 268/391 (68.5\%, ``at least as precise'') to
\textbf{341/493 (69.2\%, strictly more precise)}, and the model-level
Wilcoxon from W = 214, p = 0.010 to \textbf{W = 356, p = 1.34 × 10⁻³}.
Tables 1 and 2, the per-concept depth table (Table 3 in v1, now Table
D2), and both figures were regenerated against the corrected corpus
rather than edited.

\textbf{New in v2, not a correction to v1.} The supervised head-to-head
against logistic regression and LDA (§8.1) has no counterpart in v1. An
interim version of that experiment used a train/evaluation split that
was not pair-aware; it was re-run with a pair-aware split before
anything was quoted, and §8.1 reports only the re-run. The superseded
run is not quoted anywhere in this paper.

Full edit-level provenance is in \texttt{V2\_CHANGELIST.md} in the
accompanying repository.

\begin{center}\rule{0.5\linewidth}{0.5pt}\end{center}

\subsection{Appendix F: Common
Definitions}\label{appendix-f-common-definitions}

\emph{This appendix collects the canonical cross-paper definitions
shared by the Rosetta Program's Concept Allocation Zone (CAZ) framework
{[}Henry, 2026a{]}, Geometric Evolution Maps (GEM) (this paper), and CAZ
Validation {[}Henry, 2026c{]}. It is a compact reference for a reader
working from a single paper in isolation --- each concept's full
derivation, motivation, and supporting argument live in the paper cited
alongside it, not here.}

\textbf{CAZ --- Concept Allocation Zone.} (Developed fully in the
companion paper {[}Henry, 2026a{]} §4.)

\begin{quote}
A CAZ is a locator: a coordinate on model depth that brackets where a
concept is assembled --- not the concept itself, and not a discrete
functional module. Operationally, a CAZ is the segment of the residual
stream between two saddle points of the separation curve \(S(l)\),
around a local separation peak. The detector partitions depth into such
segments and scores each; the score grades geometric salience
(Major→Gentle) --- not causal importance, and not membership.
\end{quote}

Three governing rules: (1) \emph{No membership threshold} --- every
segment is a CAZ; ``strong'' vs.~``gentle'' is score, never a binary
cut. (2) \emph{Depth-extended, not depth-localized} --- the segmentation
tiles the full depth by construction; in the degenerate limit a single
segment can span the entire network. (3) \emph{A CAZ does not locate
causation} --- score is geometric salience, not causal importance; the
causal work can be displaced from the separation peak and can sit
outside the segment entirely.

\textbf{GEM --- Geometric Evolution Map.} (Developed fully in §3 of this
paper.)

\begin{quote}
A GEM is the trajectory record of one concept's difference-of-means
direction across one CAZ segment. The settled direction is its terminal
reading, not a co-equal component.
\end{quote}

Components: \textbf{window} --- the layer span of the CAZ segment;
\textbf{trajectory} --- the per-layer dominant directions across that
span, each estimated independently; \textbf{settled direction} --- the
trajectory's reading at the segment's final layer; \textbf{handoff
layer} (\(L_H\)) --- \(\min(\text{end}+1, N-1)\), the first layer
outside the segment, \emph{derived} from the segment boundary and never
independently detected by a rotation criterion.

\textbf{Atlas.} (Developed fully in §3 of this paper.)

\begin{quote}
A concept's atlas is the set of its GEMs in a model --- one GEM per CAZ
segment. An ordinary collection of maps: no transition structure between
them is implied. An atlas is a container, not an operational object ---
it carries no defined rule for combining its GEMs into a single
concept-level answer; any concept-level result applies its own
aggregation rule, which must be stated wherever such a result is
reported.
\end{quote}

\textbf{How they fit together.} CAZ locates; GEM charts; the handoff is
where a settled direction is first evaluated on activations it was not
estimated on, and it sits outside the segment that produced it.

\textbf{Vocabulary --- use these, not the alternatives}

\begin{longtable}[]{@{}
  >{\raggedright\arraybackslash}p{(\columnwidth - 4\tabcolsep) * \real{0.3333}}
  >{\raggedright\arraybackslash}p{(\columnwidth - 4\tabcolsep) * \real{0.3333}}
  >{\raggedright\arraybackslash}p{(\columnwidth - 4\tabcolsep) * \real{0.3333}}@{}}
\toprule\noalign{}
\begin{minipage}[b]{\linewidth}\raggedright
term
\end{minipage} & \begin{minipage}[b]{\linewidth}\raggedright
means
\end{minipage} & \begin{minipage}[b]{\linewidth}\raggedright
do not say
\end{minipage} \\
\midrule\noalign{}
\endhead
\bottomrule\noalign{}
\endlastfoot
CAZ / segment & a scored saddle-to-saddle segment; the locator & ``the
zone where the concept lives''; ``depth-localized'' \\
GEM & one segment's trajectory record & ``the GEM'' meaning a whole
concept; ``assembly event'' \\
atlas & the set of a concept's GEMs in a model & ``map'' (collides with
the M in GEM); ``manifold'' \\
settled direction & the trajectory's terminal reading, at segment's end
& ``the GEM probe''; ``the direction at \(L_H\)'' \\
handoff layer & segment.end + 1, derived & ``detected where rotation
ceases''; ``the handoff within the zone'' \\
caz\_score & geometric salience, graded & ``significance''; a membership
threshold \\
peak-distal & the comparator layer/region away from a CAZ peak &
``non-CAZ'' --- the tiling is total, so there is no non-CAZ layer in an
absolute sense \\
\end{longtable}

\emph{Full provenance and the working argument behind these definitions:
\texttt{DEFINITIONS.md} and
\texttt{papers/shared/CAZ\_DEFINITION\_OF\_RECORD.md} /
\texttt{GEM\_DEFINITION.md} in the accompanying repository.}

\begin{center}\rule{0.5\linewidth}{0.5pt}\end{center}

\subsection{References}\label{references}

\begin{itemize}
\item
  Alain, G., \& Bengio, Y. (2017). Understanding intermediate layers
  using linear classifier probes. \emph{arXiv preprint
  arXiv:1610.01644}. https://arxiv.org/abs/1610.01644
\item
  Arditi, A., Obeso, O., Syed, A., Paleka, D., Panickssery, N., Gurnee,
  W., \& Nanda, N. (2024). Refusal in language models is mediated by a
  single direction. \emph{Advances in Neural Information Processing
  Systems 37 (NeurIPS 2024)}. \emph{arXiv preprint arXiv:2406.11717}.
  https://arxiv.org/abs/2406.11717
\item
  Belinkov, Y. (2022). Probing classifiers: Promises, shortcomings, and
  advances. \emph{Computational Linguistics}, 48(1), 207--219.
\item
  Belrose, N., Ostrovsky, I., McKinney, L., Furman, Z., Smith, L.,
  Halawi, D., Biderman, S., \& Steinhardt, J. (2023). Eliciting latent
  predictions from transformers with the tuned lens. \emph{arXiv
  preprint arXiv:2303.08112}. https://arxiv.org/abs/2303.08112
\item
  Bricken, T., Templeton, A., Batson, J., Chen, B., Jermyn, A., Conerly,
  T., Turner, N., Anil, C., Denison, C., Askell, A., Lasenby, R., Wu,
  Y., Kravec, S., Schiefer, N., Maxwell, T., Joseph, N., Tamkin, A.,
  Nguyen, K., McLean, B., Burke, J. E., Hume, T., Carter, S., Henighan,
  T., \& Olah, C. (2023). Towards monosemanticity: Decomposing language
  models with dictionary learning. \emph{Transformer Circuits Thread},
  Anthropic.
  https://transformer-circuits.pub/2023/monosemantic-features/index.html
\item
  Conneau, A., Kruszewski, G., Lample, G., Barrault, L., \& Baroni, M.
  (2018). What you can cram into a single vector: Probing sentence
  embeddings for linguistic properties. In \emph{Proceedings of the 56th
  Annual Meeting of the Association for Computational Linguistics (ACL
  2018)}, 2126--2136. \emph{arXiv preprint arXiv:1805.01070}.
  https://arxiv.org/abs/1805.01070
\item
  Cunningham, H., Ewart, A., Riggs, L., Huben, R., \& Sharkey, L.
  (2023). Sparse autoencoders find highly interpretable features in
  language models. \emph{arXiv preprint arXiv:2309.08600}.
  https://arxiv.org/abs/2309.08600
\item
  Elhage, N., Hume, T., Olsson, C., Schiefer, N., Henighan, T., Kravec,
  S., Hatfield-Dodds, Z., Lasenby, R., Drain, D., Chen, C., Grosse, R.,
  McCandlish, S., Kaplan, J., Amodei, D., Wattenberg, M., \& Olah, C.
  (2022). Toy models of superposition. \emph{Transformer Circuits
  Thread}. \emph{arXiv preprint arXiv:2209.10652}.
  https://arxiv.org/abs/2209.10652
\item
  Henry, J. (2026, software). rosetta\_tools (v1.3.1). Zenodo.
  https://doi.org/10.5281/zenodo.20361433
\item
  Henry, J. (2026, software). Rosetta\_Analysis: Analysis, figure, and
  reproduction scripts for the Rosetta interpretability research program
  (v1.1.0). Zenodo. https://doi.org/10.5281/zenodo.21583139
\item
  Henry, J. (2026, software). Rosetta: Code and data for the Rosetta
  interpretability research series.
  https://github.com/jamesrahenry/Rosetta
\item
  Henry, J. (2026, dataset). Rosetta Activations: Pre-extracted
  transformer residual stream activations for 46 language models across
  17 concepts (40 with full per-model analysis; 6 frontier models
  CAZ-only). HuggingFace.
  https://huggingface.co/datasets/james-ra-henry/Rosetta-Activations
  (DOI: https://doi.org/10.57967/hf/9725)
\item
  Henry, J. (2026a). The Concept Allocation Zone: Tracking How Concepts
  Form Across Transformer Depth. \emph{arXiv preprint arXiv:2605.24856}.
  https://arxiv.org/abs/2605.24856
\item
  Henry, J. (2026c). Concept Encoding Strategies Across 28 Transformers:
  A Concept Allocation Zone and Geometric Evolution Map Evaluation.
  Zenodo. https://doi.org/10.5281/zenodo.22016917
\item
  Huh, M., Cheung, B., Wang, T., \& Isola, P. (2024). Position: The
  Platonic Representation Hypothesis. \emph{Proceedings of the
  International Conference on Machine Learning (ICML 2024)}. \emph{arXiv
  preprint arXiv:2405.07987}. https://arxiv.org/abs/2405.07987
\item
  Lieberum, T., Rajamanoharan, S., Conmy, A., Smith, L., Sonnerat, N.,
  Varma, V., Kramár, J., Dragan, A., Shah, R., \& Nanda, N. (2024).
  Gemma Scope: Open sparse autoencoders everywhere all at once on Gemma
  2. \emph{arXiv preprint arXiv:2408.05147}.
  https://arxiv.org/abs/2408.05147
\item
  Marks, S., \& Tegmark, M. (2024). The geometry of truth: Emergent
  linear structure in large language model representations of true/false
  datasets. \emph{Conference on Language Modeling (COLM 2024)}.
  \emph{arXiv preprint arXiv:2310.06824}.
  https://arxiv.org/abs/2310.06824
\item
  nostalgebraist. (2020). Interpreting GPT: The logit lens.
  \emph{LessWrong / AI Alignment Forum}, August 2020.
  https://www.lesswrong.com/posts/AcKRB8wDpdaN6v6ru/interpreting-gpt-the-logit-lens
\item
  Tenney, I., Das, D., \& Pavlick, E. (2019). BERT rediscovers the
  classical NLP pipeline. In \emph{Proceedings of the 57th Annual
  Meeting of the Association for Computational Linguistics (ACL 2019)},
  4593--4601. \emph{arXiv preprint arXiv:1905.05950}.
  https://arxiv.org/abs/1905.05950
\item
  Zou, A., Phan, L., Chen, S., Campbell, J., Guo, P., Ren, R., Pan, A.,
  Yin, X., Mazeika, M., Dombrowski, A.-K., Goel, S., Li, N., Byun, M.
  J., Wang, Z., Mallen, A., Basart, S., Koyejo, S., Song, D.,
  Fredrikson, M., Kolter, J. Z., \& Hendrycks, D. (2023). Representation
  engineering: A top-down approach to AI transparency. \emph{arXiv
  preprint arXiv:2310.01405}. https://arxiv.org/abs/2310.01405
\end{itemize}

\end{document}